\documentclass{article}

\usepackage{microtype}
\usepackage{graphicx}
\usepackage{subcaption}
\usepackage{booktabs} 

\usepackage[preprint]{icml2026}

\usepackage{amsmath}
\usepackage{amssymb}
\usepackage{mathtools}
\usepackage{amsthm}

\usepackage[textsize=tiny]{todonotes}

\usepackage[utf8]{inputenc}
\usepackage[T1]{fontenc}
\usepackage{url}
\providecommand{\Url}[1]{\url{#1}}
\providecommand{\url}[1]{\texttt{#1}}
\usepackage{booktabs}       %
\usepackage{amsfonts}       %
\usepackage{microtype}      %
\usepackage{xcolor}         %
\usepackage{amsmath}      %
\usepackage{algorithm}

\usepackage{algpseudocode}
\usepackage{amssymb}      %
\usepackage{natbib}
\usepackage{adjustbox}
\usepackage{array}
\usepackage{caption}
\usepackage{makecell}
\usepackage{epigraph}
\usepackage{float}
\usepackage{enumitem}
\usepackage{listings}
\usepackage{wrapfig}
\usepackage{xspace}
\usepackage{subcaption}
\usepackage{multirow,wasysym}
\usepackage{siunitx}

\definecolor{linkcolor}{RGB}{74, 102, 146}
\usepackage[
colorlinks=true,allcolors=linkcolor,pageanchor=true,
plainpages=false,pdfpagelabels,bookmarks,breaklinks,bookmarksnumbered,
backref=page
]{hyperref}
\usepackage[capitalize,noabbrev]{cleveref}

\newcommand{\add}[1]{\textcolor{black}{#1}}
\newcommand{\syl}[1]{\textcolor{black}{#1}}

\newcommand{\name}{BaNEL\xspace}
\definecolor{lightpurple}{RGB}{168, 141, 201}

\definecolor{darkgreen}{RGB}{0,128,0}
\newcommand{\x}{\mathbf{x}}

\newcommand{\phib}{{\boldsymbol{\phi}}}

\newcommand{\thetab}{{\boldsymbol {\theta}}}
\newcommand{\etab}{{\boldsymbol {\eta}}}

\newcommand{\beq}{\begin{equation}}
\newcommand{\eeq}{\end{equation}}
\newcommand{\beqa}{\begin{eqnarray}}
\newcommand{\eeqa}{\end{eqnarray}}

\icmltitlerunning{Exploration Posteriors for Generative Modeling using Only Negative Rewards}

\begin{document}

\twocolumn[
  \icmltitle{\name: Exploration Posteriors for \\Generative Modeling using Only Negative Rewards}



  \icmlsetsymbol{equal}{*}

  \begin{icmlauthorlist}
    \icmlauthor{Sangyun Lee}{yyy}
    \icmlauthor{Brandon Amos}{}
    \icmlauthor{Giulia Fanti}{yyy}
  \end{icmlauthorlist}

  \icmlaffiliation{yyy}{Carnegie Mellon University}

  \icmlcorrespondingauthor{Sangyun Lee}{sangyunl@andrew.cmu.edu}

  \icmlkeywords{Machine Learning, ICML}

  \vskip 0.3in
]



\printAffiliationsAndNotice{}  

\begin{abstract}
Today's generative models thrive with large amounts of supervised data and informative reward functions characterizing the quality of the generation. They work under the assumptions that the supervised data provides knowledge to pre-train the model, and the reward function provides dense information about how to further improve the generation quality and correctness. However, in the hardest instances of important problems, two challenges arise: (1) the base generative model attains a near-zero reward signal, and (2) calls to the reward oracle are expensive. This setting poses a fundamentally different learning challenge than standard reward-based post-training. To address this, we propose BaNEL (Bayesian Negative Evidence Learning), which  post-trains the model using failed attempts only, while minimizing the number of reward evaluations (NREs). \name works by learning a second, in-loop generative model for failures. We  use  this model to steer the generation away from samples that resemble previous failures. We show that \name can significantly improve model performance without observing a single successful sample on several sparse-reward tasks. It outperforms existing novelty-bonus approaches in success rate (in some cases by orders of magnitude), while using fewer reward evaluations.
\end{abstract}

\section{Introduction}

Today's generative models thrive with large amounts of supervised data
and informative reward functions characterizing the quality of the generation,
especially for generating
language, image, video, and audio.
This pipeline works well under the assumptions that
1) the supervised data provides broad enough coverage of the generation space,
and 2) the reward function provides information about how to improve
or focus the generation quality and correctness.
Language modeling with verifiable rewards~\citep{guo2025deepseek} works well because the base models often
start with at least some positive reward signal on the task.

\textbf{Challenge: Tasks with near-zero reward and expensive reward oracles.}
In many unsolved critical applications---including 
the next generation of theorem proving, algorithmic problem solving,
and drug discovery, to name a few---this standard pipeline encounters  two core challenges. (1) \emph{Sparsity:} Oftentimes, the
base generative model attains a near-zero reward signal. The probability of producing a positive-reward sample can be so low that the model may go through most of training without ever encountering one.
(2) \emph{High-cost reward evaluation:} Calls to the reward oracle can be expensive or risky, requiring costly simulations, computations, or even physical experiments \citep{korshunova2022generative}.
Hence, there is a need for \textbf{learning algorithms that can learn from exclusively negative-reward samples, while  minimizing number of reward evaluations (NREs).} 
This setting 
poses a fundamentally different learning challenge than standard reward-based post-training.
Learning in such harsh conditions is crucial: failure to tackle this challenge would mean that post-training
is merely limited to distribution sharpening rather than unlocking genuinely new capabilities.

The performance of such learning algorithms largely depends on their ability to recognize and generalize from a small number of failures; ideally, this ability should \textbf{scale with compute.}
In deep RL, reward sparsity is often addressed by introducing novelty bonuses to encourage exploration. Two of the most popular techniques for doing so include count-based methods~\citep{bellemare2016count,ostrovski2017count} and random network distillation~\citep{burda2018exploration}. These methods have proven effective in sparse-reward Atari environments such as Montezuma’s Revenge~\citep{ostrovski2017count,burda2018exploration,badia2020never,badia2020agent57}. However, quality of the intrinsic signal does not scale with compute,
and as such they must query the reward oracle frequently.
On the other hand, prediction-error approaches~\citep{schmidhuber2010formal,pathak2017curiosity,stadie2015incentivizing} learn the dynamics of the environment; these methods can be scalable but they are inapplicable for training generative models, where the transition dynamics are known and deterministic.
Recent reward-based sampling methods like GFlowNets \citep{bengio2021flow} allow for multiple parameter updates per reward evaluation, but they are unable to learn in extremely sparse environments.

\textbf{Our approach: Train a generative model on failures and update the policy distribution away from the negative samples.}
The zero-reward problem can be solved in many ways, such as using
positive transfer from other tasks or domains, hand-designing
curricula, and/or engineering more informative and dense reward functions.
We argue there will always fundamentally be tasks and settings where the base
model attains an extremely sparse reward, and that even these negative
samples provide useful information to learn and explore from.
Motivated by other sparse reward reinforcement learning methods, we propose to use the negative samples and reweight
the base distribution away from them.
Specifically, we train a generative model on negative samples for multiple epochs,
and use it to assess whether data is similar to
previously seen failures.  If a sample is similar to other zero-reward
data, the algorithm rejects it before querying the expensive reward
oracle.  This mirrors human scientists who, based on their failures,
know what is unlikely to work and thus what to try next.

In summary, we make the following contributions: 
\begin{enumerate}[leftmargin=*]
	\item \textbf{Conceptual:} We show in Section \ref{sec:explanation} why existing leading techniques for post-training generative models and learning under sparse rewards do not apply to our extremely sparse, black-box setting, where calls to the reward oracle are costly. 
	\item \textbf{Algorithmic:} We present \name (Bayesian Negative Evidence Learning), which offers three fundamental advantages for learning in extreme sparsity while minimizing calls to the reward oracle (Section \ref{sec:algo}).
    First, unlike other sparse-RL methods, it allows multiple parameter updates 	per each collected experience, allowing the model to learn efficiently from a handful of failures.  
    Second, it provides a sequential
	exploration framework that systematically narrows the search space
	until finding initial successes.
	Third, unlike many sparse RL methods, 
    \name is based on Bayesian updates which
	modify the prior multiplicatively and never explicitly decrease the
	model's likelihood for failed attempts, better preserving the model's 
	pre-trained knowledge.  
	\item \textbf{Evaluation:} We propose new experimental settings that enable controlled testing of exploration strategies for post-training generative models under sparse-reward conditions. We evaluate \name in these sparse environments and tasks in Section \ref{sec:eval}. On these experiments, \name achieves a success rate up to several orders of magnitude higher than existing baselines for the same NRE budget. We also demonstrate how it can be used to guide human intuition for solving challenging problems; in one case study, we combine \name with human intuition to learn a strategy for generating adversarial examples with 0.99 success rate, while the prior success rate was only $3\times 10^{-4}$.
    Finally, we show that \name enables trading off computation for success rate, in a new form of compute scaling. 
    
\end{enumerate}

\section{Problem Formulation: Efficient Learning from Sparse Rewards}

Let $\mathcal V$ be the discrete token set and $\mathcal V^{*}$ be the set of all finite strings over $\mathcal V$.
Define the probability distribution of our \textbf{pre-trained generative model} as $p_{\thetab}: \mathcal V^{*} \to [0, 1]$ satisfying $\sum\limits_{\x \in V^{*}} p_{\thetab}(\x)=1$ with parameter $\thetab$.
We further assume a given, binary \textbf{reward function} $r: \mathcal V^{*} \to \{0, 1\}$, where 1 and 0 mean success and failure, respectively.
The success rate of the model $\rho(p_{\thetab})$ is defined as $\rho(p_{\thetab}):= \sum\limits_{\x} p_{\thetab}(\x)r(\x)$.

The goal of reward-based training is to further improve $\rho(p_\thetab)$ 
without any additional supervised data.
In particular, we assume that evaluating $r$ is costly or risky---for instance, this can occur when running clinical trials in drug development, performing large-scale simulations \citep{korshunova2022generative}, or other cases involving direct interaction with the real world.

\noindent \paragraph{Problem Statement.} Consider a pre-trained $p_\thetab$ with a success rate $\rho(p_\thetab)$ that is so low that the model \emph{does not encounter positive examples} during training with high probability.
Our goal is to find a new model $p_\etab$ parameterized by $\etab$  such that success rate $\rho(p_\etab) \gg \rho(p_\thetab)$, while minimizing the number of calls to the reward oracle $r$, which we denote number of reward evaluations (NREs). 

Note  that we are \emph{not} necessarily trying to minimize overall computation---we want to minimize NREs, but we are willing to scale (increase) compute to make better use of reward-labeled samples.

\section{Existing Methods Fail to Address Extreme Reward Sparsity} 
\label{sec:explanation}

Our problem formulation calls for two properties:
\begin{enumerate}[leftmargin=*]
    \item \textbf{Functionality:} The algorithm should increase the prior success rate, i.e., it results in $\rho(p_\etab) \gg \rho(p_\thetab)$, given enough calls to the reward oracle.
    \item \textbf{Low number of reward evaluations (NRE):} The algorithm should efficiently use the  reward oracle $r$, e.g.,  by running multiple learning iterations  per reward evaluation. 
\end{enumerate}

\begin{table}[t]
\centering
\resizebox{\columnwidth}{!}{%
\begin{tabular}{|cc|c|c|}
\hline
\multicolumn{2}{|c|}{\textbf{Method}}                                  & \textbf{Functionality}     & \textbf{Low NREs}      \\ \hline
\multicolumn{1}{|c|}{\multirow{2}{*}{Policy Gradient}} & Classic                 & \Circle & \Circle \\ \cline{2-4}
\multicolumn{1}{|c|}{}                           & Negative RL & \Circle & \Circle \\ \hline
\multicolumn{1}{|c|}{\multirow{2}{*}{Intrinsic Rewards}} & RND                 & \LEFTcircle & \Circle \\ \cline{2-4}
\multicolumn{1}{|c|}{}                           & Count-based methods & \LEFTcircle & \Circle \\ \hline
\multicolumn{2}{|c|}{GFlowNets}                                        & \Circle     & \CIRCLE \\ \hline
\multicolumn{2}{|c|}{BaNEL (Ours)}                                     & \CIRCLE     & \CIRCLE \\ \hline
\end{tabular}%
}
\caption{Comparison of desired properties from Section \ref{sec:explanation}---functionality and low number of reward evaluations (NREs)---for key categories of learning methods. An empty circle $\Circle$ means the property is not satisfied, a filled circle $\CIRCLE$ means satisfied, and a half-filled circle $\LEFTcircle$ means partially satisfied (e.g., a method is functional, but success rate does not increase much).}
\vspace{-5mm}
\label{tab:comparison}
\end{table}

We consider several categories of algorithms with respect to our problem requirements. Our high-level assessment of these methods is included in Table \ref{tab:comparison}, with a more in-depth explanation below. 
Additional related work can be found in Appendix \ref{sec:related-extended}.

\subsection{Warm-Up Example: Policy Gradient}
We start with the well-known
policy gradient~\citep{williams1992reinforce},  the most common approach for post-training generative models from reward functions.
It has achieved great success in challenging real-world tasks, including code synthesis and math problem solving~\citep{guo2025deepseek}.

\paragraph{Classic policy gradient: zero rewards produce zero gradient}
Under classic policy gradient, we draw $m$ samples $(\x_1,\ldots,\x_m)$, where $\x_i\sim p_\thetab$. If all of these samples receive zero reward, the standard REINFORCE policy gradient is zero:
$
    \frac{1}{m} \sum_{i=1}^m r(\x_i) \nabla_\thetab \log p_\thetab(\x) = 0.
$
In this setting, policy gradient becomes brute-force random sampling until discovering the first rare success. 
By definition, this cannot improve success rate over $\rho(p_\thetab)$. 
Moreover, we cannot update our model more than once per reward evaluation without resorting to other off-policy learning techniques. 

\paragraph{Negative RL}
A straightforward way to enable learning is to subtract a constant baseline of 1:
\begin{align}
    \sum_{i=1}^m \left(r(\x_i) - 1\right) \nabla_\thetab \log p_\thetab(\x_i) = -\sum_{i=1}^m \nabla_\thetab \log p_\thetab(\x_i),
    \label{eq:negative-pg}
\end{align}
thereby suppressing model likelihood on poor samples.
Although the expected gradient remains zero, due to the finiteness of $m$, this now produces nonzero empirical gradients that we can now use for training.
\citep{zhu2025surprising} shows that incorporating negative RL along with positive examples can be beneficial in LLM training.
However, training exclusively on negative examples for an extended period breaks the model's pre-trained knowledge, leading to catastrophic collapse and rendering the model unusable for most tasks.
See Fig.~\ref{fig:neg_rl} in appendix.

\subsection{Sparse RL Techniques: Intrinsic Rewards}
\label{sec:sparse}
In response to these well-known challenges, there is a vast literature on RL under sparse rewards. For our purposes, two relevant categories of algorithms can handle all-negative-reward samples in the context of post-training a generative model.

\paragraph{Count-based methods} Count based methods introduce an exploration bonus based on state visitiation counts to reward novelty~\citep{bellemare2016count,ostrovski2017count}. Modern pseudo-count approaches~\citep{ostrovski2017count} employ a neural density model $\rho$ to approximate state visitation.
Given an observation $\x$, the density model is updated once to yield a new model $\rho'$, and the intrinsic reward is defined as some increasing function of  $\log \rho'(\x) - \log \rho(\x)$.
Count-based methods do not naturally support conducting multiple updates per reward evaluation; the density model is updated only once~\citep{bellemare2016count,ostrovski2017count}.
Applying multiple updates would artificially inflate $\log \rho'(\x) - \log \rho(\x)$, producing large bonuses even for non-novel states.

\paragraph{Random Network Distillation (RND)} 
RND instead encourages exploration by training two separate networks sharing the same architecture---a \emph{target} network, which is randomly initialized to produce an embedding of an input sample, and a \emph{predictor} network, which is trained to reduce MSE with the target network \citep{burda2018exploration}. 
The MSE between the target and the predictor is used as a curiosity bonus; when  the predictor does not match the target network, it suggests an unfamiliar state, leading to a higher MSE (and exploration bonus). 
RND can also be used to post-train LLMs~\citep{gao2025navigate}.
This method is particularly good for exploring sparse-reward regimes, but like count-based methods, it does not inherently allow for multiple updates per reward evaluation; doing so will decrease the MSE regardless of whether $\x$ is novel or not.
This can increase its NREs (Section \ref{sec:eval}).

\subsection{Reward-Based Sampling: GFlowNet}

GFlowNet~\citep{bengio2021flow} is designed to sample from a given reward function. Unlike policy gradient and most intrinsic motivation methods, it naturally supports multiple parameter updates per reward evaluation.
The most common  training objective for  GFlowNet is the Trajectory Balance loss  $\mathcal L_{TB}$ due to
\citet{malkin2022trajectory}:
\begin{align}
    \mathcal L_{TB} (\thetab, \hat Z) &:= \frac{1}{m} \sum_{i=1}^m \left(\log p_\thetab (\x_i) - \log \frac{r(\x_i) + \epsilon}{\hat Z}\right)^2 \nonumber \\
    &= \frac{1}{m} \sum_{i=1}^m \left(\log p_\thetab (\x_i) - \log \frac{\epsilon}{\hat Z}\right)^2
\end{align}
where $\hat Z$ is a free learnable parameter jointly optimized along with $\thetab$, and $\epsilon$ is a small constant to make sure the loss is defined even when $r(\x_i)=0$.
One can fix $\thetab$ and solve for $\hat Z$ to get the batch-optimal $\hat Z$ in a closed form, resulting in the VarGrad-fashion loss function~\citep{richter2020vargrad}:
\begin{align}
\mathcal L_{TB_{Vargrad}}\!(\thetab) &:= \frac{1}{m} \sum_{i=1}^m \Biggl(\log p_\thetab (\x_i) \nonumber\\
&\qquad\qquad\quad - \frac{1}{m} \sum_{j=1}^m \log p_\thetab(\x_j)\Biggr)^{\!2}\!.
\label{eq:tb_vargrad}
\end{align}
As shown above, the trajectory balance loss becomes the empirical variance of $\log p_\thetab(\x)$ over $m$ samples, so the optimal $p_\thetab$ assigns an arbitrary constant mass over $m$ samples; the remaining probability mass is distributed uncontrollably.
Hence, in the extremely sparse setting, GFlowNet fundamentally cannot learn; the resulting detachment is shown empirically in Figure \ref{fig:combined}.

\section{Avoiding Failures with Bayesian Negative Evidence Learning}
\label{sec:algo}

We now present \name (Bayesian Negative Evidence Learning). 
Our aim is to improve the policy's success rate using only reward zero experiences, without any problem-specific surrogate objectives.

\textbf{Naive idea.}
If our budget for evaluating $r$ were unlimited, we could trivially achieve a perfect success rate by collecting every possible mistake $R:=\{\x \in \mathcal V^* \mid r(\x)=0\}$ and avoiding all elements of $R$:
\begin{align}
    p_{\thetab \mid R^C}(\x) \propto p_\thetab(\x) \mathbf 1[\x \notin R].
\end{align}
Here, $\mathbf 1 [\cdot]$ denotes the indicator function, and we define $p_{\thetab \mid S}(\x):=\frac{p_{\thetab}(\x) \mathbf 1[\x \in S]}{\sum_{\x} p_{\thetab}(\x) \mathbf 1[\x \in S]}$ given a set $S$.
We use $S^C$ to denote the complement in $\mathcal V^*$ of a set $S$.
This approach is infeasible because the space of failures is combinatorial and we want to minimize NREs.
Fortunately, in most tasks, failures exhibit underlying regularities. In such cases, a neural network can learn to recognize and generalize from these patterns, removing the need to encounter every instance.
Thus, the key factor determining performance is the model’s ability to infer the failure set $R$ from only a limited number of examples.
Ideally, we want this ability to scale with compute.

\subsection{Learning a Generative Model for Failed (Zero-Reward) Attempts}

We cast the problem of learning regularities in failures  as another, in-loop generative modeling problem.
Specifically, we train a separate likelihood-based generative model $p_\phib$ (parameterized by $\phib$) on $m$ negative examples with the standard maximum likelihood objective:
\begin{align*}
\max_\phib \frac{1}{m} \sum_{i=1}^m \log p_\phib (\x_i).
\end{align*}
Once well-trained, $p_\phib(\x)$ can be used to assess whether a given input resembles previously observed failures;
specifically, we use $p_\phib$ to define a rejection region $\tilde R$ approximating $R$.

For that, the rejection region $\tilde R$ should contain samples that are likely for $p_\phib(\x)$ so the model can avoid making similar mistakes to previously-made ones.
To this end, we define $\tilde R$ as follows:
\begin{align}
    \tilde R:= \left\{\x : \frac{p_\thetab(\x)}{p_\phib(\x)} < \tau \right\}
    \label{eq:rejection-set-definition}
\end{align}
where $\tau$ is a (potentially data-dependent) threshold value.~\footnote{\syl{We use likelihood ratio instead of raw likelihood following out-of-distribution detection literature. See \citep{nalisnick2018deep,ren2019likelihood}.}}
Note that this requires $p_\thetab$ and $p_\phib$ to be  likelihood-based generative models under which we can compute the likelihood.
Using the rejection region $\tilde R$, we form a Bayesian posterior $\tilde p_\thetab$ to approximate $p_{\thetab \mid R^C}$:
\begin{align}
    p_{\thetab \mid \tilde R^C}(\x) \propto p_\thetab(\x) \mathbf{1} [\x \notin \tilde R]~,
    \label{eq:posterior-definition}
\end{align}

This policy filters out data points that are similar to prior failures according to $\tilde R$;  equivalently, we direct the model to sample only from $\tilde R^C$.

\textbf{Success rate analysis.}
Recall that success rate is defined as $\rho(p):=\sum\limits_{\x}p(\x)r(\x)$.
The success rate of the posterior can be written as follows:
\begin{align*}
\rho(p_{\thetab \mid \tilde R^C}) &= \sum_{\x \in \tilde R^{C}} p_{\thetab \mid \tilde R^C} (\x) r(\x) \\
&= \sum_{\x \in \tilde R^{C}} \frac{p_{\thetab}(\x \in \tilde R^{C}|\x)p_{\thetab}(\x)}{p_{\thetab}(\tilde R^{C})} r(\x) \\
&= \frac{1}{p_{\thetab}(\tilde R^{C})} \sum_{\x \in \tilde R^{C}} p_{\thetab}(\x)  r(\x)\\
&= \frac{1}{1-p_{\thetab}(\tilde R)} \left(\rho(p_{\thetab}) - \sum_{\x \in \tilde R} p_{\thetab}(\x)  r(\x)\right)\\
&= \frac{\rho(p_{\thetab})}{1-p_{\thetab}(\tilde R)} - \frac{p_\thetab(\tilde R)}{1-p_{\thetab}(\tilde R)} \rho(p_{\thetab \mid \tilde R}),
\end{align*} 
where we abuse notation to denote $p_\thetab(S)=\sum_{s \in S} p_\thetab(s)$ for some set $S$.
The above decomposition gives qualitative insights about the desired properties of $\tilde R$:
\begin{itemize}[leftmargin=*]
    \item \textbf{Misclassification rate of $\tilde R$.} The posterior success rate decreases when $\rho(p_{\thetab \mid \tilde R})$  increases, so we need to train $p_\phib$ well and define $\tilde R$ properly so that $\tilde R$ does not misclassify $r=1$ samples and mistakenly reject them.
    \item \textbf{Make $\tilde R$ as large as possible.} If we can drive $\rho(p_{\thetab \mid \tilde R})$ close to zero, the posterior success rate is roughly $\frac{1}{1-p_{\thetab}(\tilde R)}$ times greater than the prior and approaches 1 as $\tilde R$ grows.
\end{itemize}
Nevertheless, $\tilde R$ does not need to be perfect, as  $\rho(p_{\thetab \mid \tilde R}) \leq \rho(p_\thetab) \implies \rho(p_{\thetab \mid \tilde R^C}) \geq \rho(p_{\thetab})$.

\begin{algorithm}[t]
\caption{Sequential Filtering (No Distillation)  }
\label{alg:seq-nodistill}
\footnotesize
\begin{algorithmic}[1]
\State \textbf{Initialize} iterations \(n\).
\State Sample \(\{\x_j\}_{j=1}^{m} \sim p_\thetab^{}\).
\State \textbf{Fit failure model} \(p_{{\phib^{0}}}(\x)\) by maximizing \(\frac{1}{m}\sum_{j=1}^{m}\log p_{\phib^{0}}(\x_j)\).
\For{\(i = 1\) \textbf{to} \(n-1\)}
  \State Sample $\{\x_j\}_{j=1}^{m}$ from
  $p_\thetab (\x) \prod_{k=0}^{i-1} \mathbf{1} \!\bigl[ \tfrac{p_\thetab(\x)}{p_{\phib^{k}}(\x)} \geq \tau \bigr]$
  \State Evaluate $\{r(\x_j)\}_{j=1}^{m}$. Terminate if $r(\x_j) = 1$ for any $j$.
  \State \textbf{Fit failure model} \(p_{\phib^{i}}(\x)\) by maximizing \(\frac{1}{m}\sum_{j=1}^{m}\log p_{\phib^{i}}(\x_j)\).
\EndFor
\State \Return $p_\thetab (\x) \prod_{k=0}^{n-1} \mathbf{1} \!\bigl[ \tfrac{p_\thetab(\x)}{p_{\phib^{k}}(\x)} \geq \tau \bigr]$.
\end{algorithmic}
\end{algorithm}
\vspace{-5mm}

\paragraph{Adaptive selection of  rejection region $\tilde R$}
As the rejection threshold $\tau$ increases,  so does $p_\thetab (R)$, and hence $\tilde R$ rejects samples more aggressively.
However, the same threshold $\tau$ could result in drastically different rejection regions $\tilde R$ for different negative-sample models $p_\phib$. 
To simplify design, we adaptively choose $\tau$ so that
we accept a fixed number of $m$ samples in each batch.
To generate $m$ samples, we first draw $mf$ samples from the prior, for some filtering factor $f>1$.
We then sort the $mf$ samples in descending order of likelihood ratio $\frac{p_\thetab(\x)}{p_\phib(\x)}$, and only accept the first $m$ samples.
$f=1$ means $\tilde R$ is empty, whereas a larger $f$ indicates that only samples that are much more likely in our prior $p_\thetab$ than in our negative model $p_\phib$ are accepted.

\textbf{Relationship with Cross Entropy Method (CEM).}
When $\tau$ is chosen adaptively so that exactly $m$ of the $mf$ candidates are accepted, the procedure coincides with the elite-selection step of the cross-entropy method (CEM)~\citep{de2005tutorial}.
The key difference is that CEM ranks candidates by reward, whereas in our setting reward is always zero, so we instead use the likelihood ratio $\frac{p_{\thetab}(\x)}{p_{\phib}(\x)}$ as a surrogate ranker.
As a soft alternative, we also tried importance resampling with weights proportional to this likelihood ratio (analogous to replacing CEM's hard cut with weights), but it did not yield consistent improvements.
For simplicity, we therefore adopt the CEM-style hard cut.

\subsection{Combining Multiple Filters Efficiently via Distillation}

The proposal distribution can be refined online by repeating Bayesian updates as new samples arrive. In this sequential approach, rejection regions from earlier rounds can be accumulated by taking their union (i.e., $\tilde R \gets \tilde R \cup \tilde R_{\text{new}}$ where $R_{\text{new}}$ is the new rejection region). 
This yields Algorithm~\ref{alg:seq-nodistill}.\footnote{We omit the partition function of the unnormalized distributions to simplify notation from now on.}

\begin{figure*}[t]
    \centering
    \includegraphics[width=\textwidth]{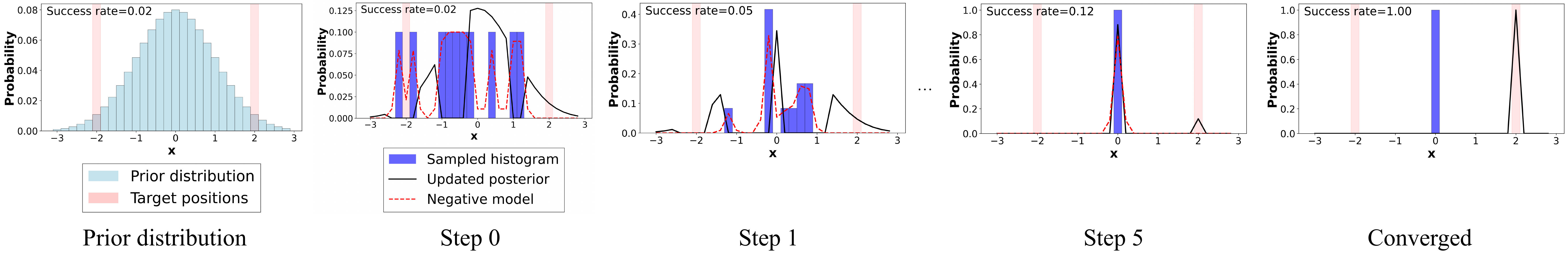}
    \caption{Illustration of \name on a 1D toy example with negative-reward samples only. The procedure begins with a pre-trained proposal distribution (leftmost). Two reward-one \syl{positions} (red bars) are located at -2 and 2. At each iteration, the proposal distribution generates samples, which are very likely to be 0-reward, \syl{as indicated by low success rate in top-left corner.} These are used to train a negative model (red dashed curves). The proposal and negative models are combined to form the Bayesian posterior (black curves), following Eq.~\eqref{eq:posterior-definition}. As iterations progress, the posterior increasingly concentrates on the reward-one regions, until convergence (rightmost).
    }
    \label{fig:1d}
\end{figure*}

However, this algorithm is not practical because of two reasons: (1) it requires maintaining multiple negative models for filtering, and (2) since the prior rarely generates samples outside all the rejection regions, rejection sampling can become very inefficient.
We handle this issue by distilling the filtered distribution into the model at each stage, leading to Algorithm~\ref{alg:seq-distill} (main difference highlighted in \textcolor{blue}{blue}).
In practice, we implement the distillation step  via maximum likelihood training, reusing the same $m$ samples to train the failure model for efficiency. 
\syl{\textbf{As Algorithm~\ref{alg:seq-distill} is theoretically equivalent to Algorithm~\ref{alg:seq-nodistill} while being significantly more efficient, we use Algorithm~\ref{alg:seq-distill} in our experiments.}}
See Fig.~\ref{fig:1d} for a visual illustration of the algorithm.

\begin{algorithm}[t]
\caption{Sequential Filtering with Distillation }
\label{alg:seq-distill}
\footnotesize
\begin{algorithmic}[1]
\State \textbf{Initialize} \(p_{\thetab^{0}}(\x) \gets p_\thetab\); iterations \(n\)
\State Sample \(\{\x_j\}_{j=1}^{m} \sim p_{\thetab^{0}}\).
\State \textbf{Fit failure model} \(p_{\phib^{0}}(\x)\) by maximizing \(\frac{1}{m}\sum_{j=1}^{m}\log p_{\phib^{0}}(\x_j)\).
\For{\(i = 1\) \textbf{to} \(n-1\)}
  \State Sample \(\{\x_j\}_{j=1}^{m} \sim p_{\thetab^{i-1}}(\x) \mathbf{1} \!\bigl[ \tfrac{p_\thetab(\x)}{p_{\phib^{i-1}}(\x)} \geq \tau \bigr]\) .
  \State Evaluate $\{r(\x_j)\}_{j=1}^{m}$. Terminate if $r(\x_j) = 1$ for any $j$.
  \State \textbf{Fit failure model} \(p_{\phib^{i}}(\x)\) by maximizing \(\frac{1}{m}\sum_{j=1}^{m}\log p_{\phib^{i}}(\x_j)\).
  \State \textcolor{blue}{\textbf{Distill} the filter into the model:
        \(p_{\thetab^{i}}(\x) \gets p_{\thetab^{i-1}}(\x) \mathbf{1} \!\bigl[ \tfrac{p_\thetab(\x)}{p_{\phib^{i-1}}(\x)} \geq \tau \bigr]\).}
\EndFor
\State $\etab \gets \thetab^{n-1}$
\State \Return \(p_{\etab}\).
\end{algorithmic}
\end{algorithm}
\vspace{-5mm}

\section{Experiments}
\label{sec:eval}
We evaluate \name by constructing new sequential generation tasks with extremely sparse rewards. 
In Sec.~\ref{exp-mnist}, we evaluate on MNIST~\citep{lecun1998mnist}, where we can visualize exploration.
\syl{In Sec.~\ref{sec:exp-attack}, we consider adversarial attack scenario against a target language model.}
In Sec.~\ref{exp-gsm}, we  test on a challenging subset of GSM8K~\citep{cobbe2021training} reasoning tasks where pretrained models fail. 
In these experiments, we \textbf{deliberately filter out reward one samples} to test an algorithm’s ability to learn from zero-reward observations only.
We compare \name (ours) to the random network distillation \citep{burda2018exploration} and pseudo-count based methods \citep{ostrovski2017count} baselines.
In Sec.~\ref{sec:exp-attack} we provide extra results where the attacker generates digit-addition problems that the target model misanswers.
Appendix~\ref{sec:ablation} includes ablations that show the effect of various hyperparameters and other design choices regarding the distillation step in Algorithm \ref{alg:seq-distill}.

\begin{figure}[]
    \centering
    \includegraphics[width=0.65\linewidth]{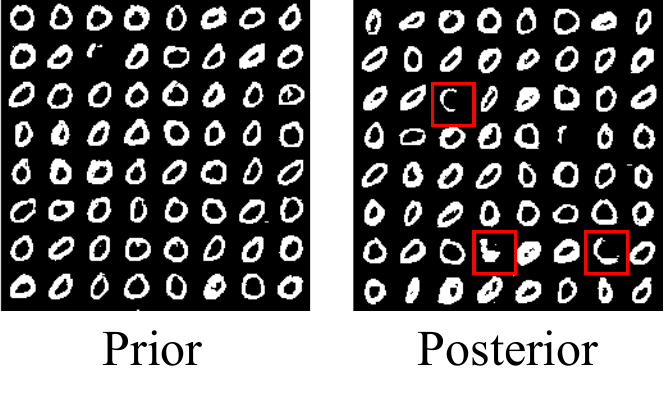}
    \caption{Prior samples (left, success rate: 8e-26) and the best posterior samples from our method (right, success rate: 5e-21).
    }
    \label{fig:mnist}
    \vspace{-3mm}
\end{figure}

\subsection{MNIST $0 \rightarrow 6$}
\label{exp-mnist}
In this task, we pre-train autoregressive generative models on the 0-digit subset of the MNIST training set, and the task is to discover 6's.
Since a 0 is visually close to a 6 digit, pre-training increases the success rate significantly. 
At the same time, a 6 can only be discovered by doing a significant exploration from 0, testing the algorithm's ability to generate new knowledge.

To summarize our setting: 
    Our pre-trained model $p_\thetab$ is an autoregressive transformer trained on 0 digits.
    Our reward $r(\x)=1$  if the model generates data exactly matching any element of the \emph{target set}, a set of 50,000 6-digits generated by applying random affine transformations to the MNIST 6-digits in the test set.
    This experimental setting has \emph{extreme reward sparsity.} The base model's success rate is 8e-26.~\footnote{\syl{We can compute $\log p_\thetab(\x)$ explicitly as $p_\thetab$ is an autoregressive generative model whose likelihood is tractable. We compute success rate $\sum_\x p_\thetab(\x)$ by exponentiating and summing up the log likelihood values using $\mathrm{torch.logsumexp()}$.}}
    We set the total NRE budget to 7500 for all methods.

\begin{figure}[]
    \centering
    \includegraphics[width=0.8\linewidth]{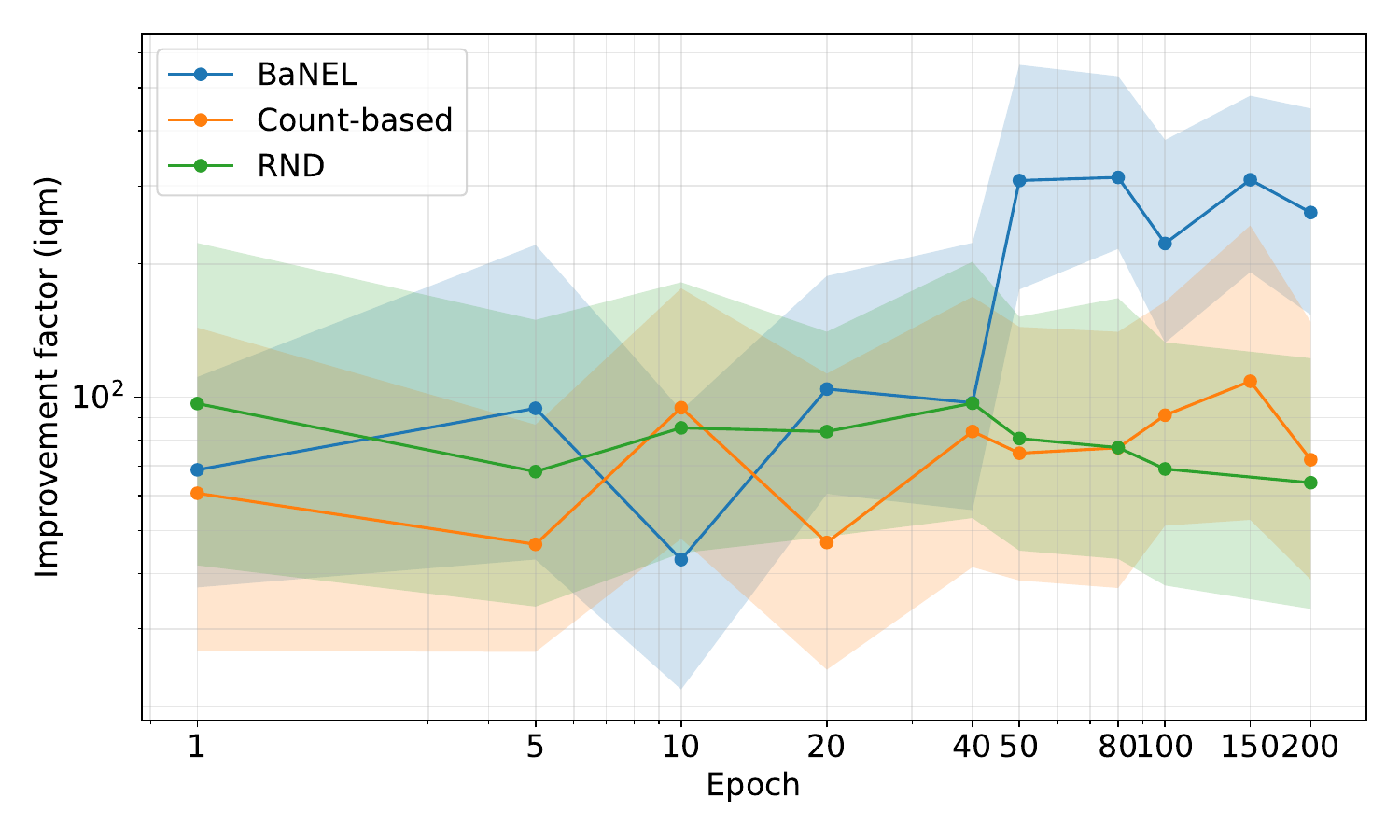}
    \caption{\add{Compute scaling: Interquartile mean (IQM; mean of the middle 50\%~\citep{agarwal2021deep}) improvement factor in success rate over the base model for BaNEL, count-based, and RND as a function of the number of training epochs. For BaNEL, the x-axis is the number of epochs used to train $p_{\phib}$ at each stage; for RND and count-based methods, it is the number of epochs used to train the random network and density model per rollout. IQM values are computed over \textbf{100 random seeds.} Shaded regions indicate 95\% bootstrap confidence intervals.}}
    \label{fig:mnist_actor_neg_epoch}
\end{figure}

\begin{figure*}[t]
\centering

\begin{subfigure}[b]{.33\textwidth}
  \centering
  \includegraphics[width=\linewidth,height=0.75\linewidth,keepaspectratio]{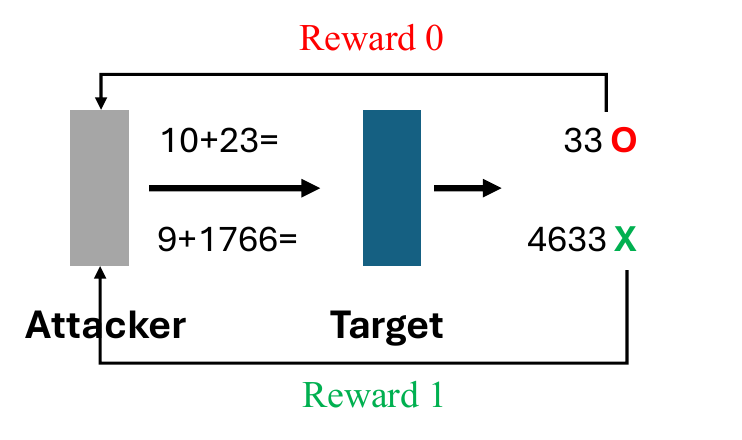}
  \caption{}\label{fig:a}
\end{subfigure}\hfill
\begin{subfigure}[b]{.33\textwidth}
\begin{lstlisting}[
  basicstyle=\ttfamily\small\color{black},
  frame=single, framerule=0.5pt, rulecolor=\color{darkgreen},
  xleftmargin=0pt, framexleftmargin=0pt, framexrightmargin=0pt,
  aboveskip=0pt, belowskip=0pt
]
# Leading zeros
000840040+6336084=
04967+660843=
006509+602096=

# Carry-chain stressors
4057539400+6460920=
5108069997+50003=
99999999+9=
\end{lstlisting}
\caption{}\label{fig:b}
\end{subfigure}\hfill
\begin{subfigure}[b]{.30\textwidth}
\centering
\footnotesize
\begin{tabular}{@{}l S[table-format=2.2]@{}}
\toprule
Pattern & {Rate (\%)} \\
\midrule
Pre-trained  & 0.04 \\
Carry chain  & 99.02 \\
Leading zeros & 99.96 \\
\bottomrule
\end{tabular}
\caption{}\label{fig:c}
\end{subfigure}

\caption{(a) Adversarial attack setup for Sec.~\ref{sec:exp-attack}. {If a prompt returns a correct answer, it has zero reward (red 'o'); incorrect answers have reward one (green 'x')}; (b) examples of successful attacks found by \name; (c) rule-based attack results using patterns in (b).}
\label{fig:adv-combined}
\end{figure*}

\paragraph{\name's success rate scales with compute}
Unlike prior sparse RL techniques, \name can utilize  additional compute to improve its success rate, even for a fixed number of NREs.
{Fig.~\ref{fig:mnist_actor_neg_epoch} shows that the performance of \name tends to increase as the number of epochs used to train $p_\phib$ at each stage increases} unlike other two methods. This indicates that while the benefit of \name becomes effective when additional computation is available to extract richer knowledge from failures (unlike our baselines, which cannot exploit additional computation).

Fig.~\ref{fig:mnist} shows that, in the posterior samples, digits shaped like a '0' with the right side removed--thereby resembling a '6'--occur more frequently than in the prior.

\subsection{\add{Adversarial Attack on Toy Language Model}}
\label{sec:exp-attack}

\begin{table}[]
\vspace{-3mm}
\footnotesize
\caption{\add{Best interquartile mean (IQM; mean of the middle 50\%~\citep{agarwal2021deep}) improvement factor in success rate over the base model for BaNEL, count-based, and RND. The upper and lower values of 95\% bootstrap confidence intervals are also reported. Pre-trained model's success rate is roughly 0.0004. IQM values are computed over \textbf{100 random seeds.}}}
\centering
\begin{tabular}{lccc}
\toprule
Method & IQM & CI lower & CI upper \\
\midrule
Ours        & \textbf{48.51$\times$} & \textbf{34.74$\times$} & \textbf{66.93$\times$} \\
Count-based & 1.62$\times$  & 1.56$\times$  & 1.67$\times$  \\
RND         & 1.45$\times$  & 1.43$\times$  & 1.47$\times$  \\
\bottomrule
\end{tabular}
\label{tab:adv}
\end{table}

In this task, the goal is to attack the \textit{target model}, an autoregressive transformer trained to answer digit-addition queries (e.g., it receives \texttt{10+23=} and must generate \texttt{33}). The goal of the \textit{attacker model}, also an autoregressive transformer trained to generate questions such as \texttt{10+23=}, is to propose syntactically valid addition queries on which the target model produces an incorrect sum. Both models use the GPT-2 architecture (we use \href{https://github.com/karpathy/nanoGPT}{nanoGPT}) with a character-level tokenizer; the vocabulary comprises the ten digits $\{0,\dots,9\}$, arithmetic symbols (e.g., \texttt{+}, \texttt{=}), and alphabetic characters.
The maximum length of each operand is set to 10.
We define the reward as follows:

$$
r(\x)\!=\!
\begin{cases}
1, & \parbox[t]{0.58\columnwidth}{if $\x$ is a syntactically valid arithmetic expression and the target's output is incorrect,}\\[4pt]
0, & \text{otherwise,}
\end{cases}
$$

and the target is evaluated using greedy decoding. Because grammatically invalid sequences receive zero reward by construction, pre-training the attacker on the same distribution of digit-addition problems is necessary so that it reliably proposes syntactically valid expressions that the target can parse and attempt to answer. 
See Fig.~\ref{fig:adv-combined}(a) for visual explanation.
Since the target is trained well, the pre-trained attacker's empirical success rate is roughly $0.0004$ (Clopper-Pearson CI: $[0.00032,\,0.00047]$;  num\_samples$=300{,}000$, $\alpha=0.05$).

\add{Table~\ref{tab:adv} shows that BaNEL outperforms other methods by a large margin.}
In addition to increasing the raw success rate, this experiment surfaced several qualitative patterns. Fig.~\ref{fig:adv-combined} (b) shows two examples of successful attacks.
\name identifies two failure modes of the target: \textit{(1) Leading zeros}: when at least one of the input digits start with at least one zero, the output result tends to be incorrect.  
Note that {the attacker model had never seen leading zeros during pre-training.}
\textit{(2) Carry-chain stressors} refer to examples that need to carry a digit during summation. 
Together, these two failure classes explain a large fraction of successful attacks found by \name. 

Based on the insights discovered by \name, we write a script to generate questions following these two patterns to attack the target model.
Specifically, we generate 5120 samples from each pattern, and compute the resulting success rate. 
Fig.~\ref{fig:adv-combined}(c) shows that the final success rate is near 1.
This suggests that \name can be used both to increase a numeric success rate, but it can also be useful to guide human intuition on hard problems to extract qualitative insights.
See Appendix~\ref{sec:imp_detail} for more details on how the rule-based attacks are generated.
For completeness, we provide additional results where we do not allow for leading zero attacks (Appendix~\ref{exp:adv-additional}).

\subsection{GSM8K-Hard}
\label{exp-gsm}
Next, we compare \name with RND (following the implementation of \citet{gao2025navigate}), the strongest baseline on MNIST setting, on GSM8K-Hard, a challenging subset of GSM8K dataset~\citep{cobbe2021training}.
\syl{We first fine-tune Qwen 2.5 0.5B Instruct model~\citep{qwen2025qwen25technicalreport} with PPO on GSM8K train set until convergence.
We then evaluate the model on GSM8K test set and select problems where the model attains a success rate between $1\times10^{-4}$ and $3\times 10^{-3}$.
This range is small enough to reflect the challenge of sparsity, yet not so small that empirical estimation of success rates becomes impractical.
Specifically, we choose the following question IDs: 143, 1248, 1012, 510, 942, and 205.
We treat each of these questions as a separate sparse-reward environment.
We set the NRE budget to 7680.}

\syl{GSM8K-Hard is different from the standard setup for RL for LLM. Here, \textbf{there is no distinction between train and test set, or in other words, we directly train on test set.} Moreover, we filter out any positive reward signals, so \textbf{the model should learn from zero-reward samples only, making this task extremely challenging.}
}

\begin{figure}
    \centering
    \includegraphics[width=1.\linewidth]{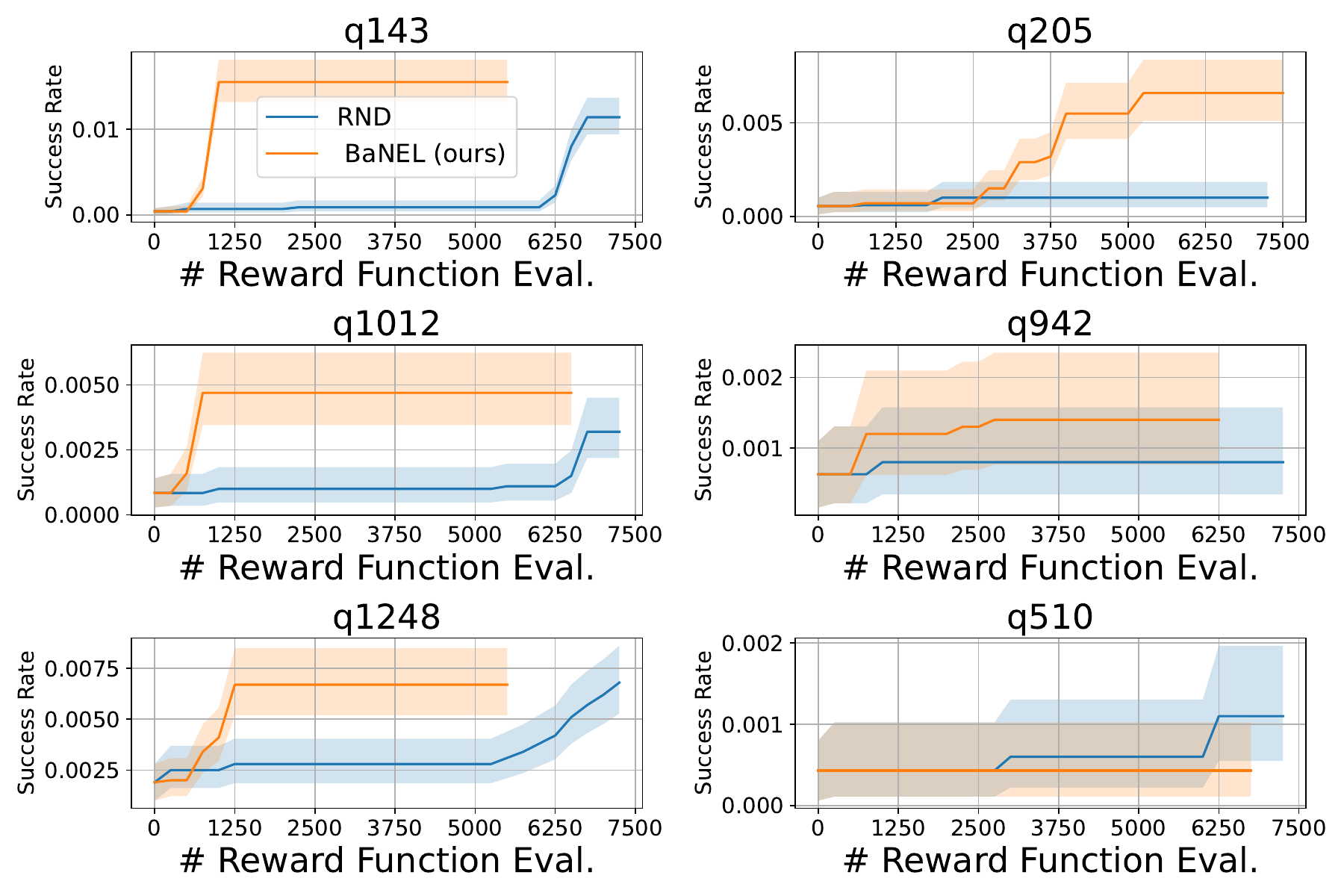}
    \caption{Cumulative best success rate of BaNEL and RND on GSM8K-Hard questions. Shaded area represents confidence intervals (Clopper-Pearson, $\alpha=0.05$, sample\_size=10000).}
    \label{fig:gsm}
    \vspace{-5mm}
\end{figure}

As shown in Fig.~\ref{fig:gsm}, \textbf{\name strictly outperforms RND on 4 problems (143, 205, 1012, and 942)}, achieving higher success rates with significantly fewer NRE. On one problem (1248), \name achieves a comparable success rate while requiring roughly $6\times$ fewer NREs, and on the remaining problem (510), RND outperforms \name.
These results demonstrate that \name learns and generalizes more effectively than RND from failure-only feedback.
Note that Fig. \ref{fig:gsm} shows the historical maximum success rate of each baseline. This is an appropriate visualization because the NREs are only an upper bound; in practice, one can always use fewer. The raw values are plotted in Fig.~\ref{fig:gsm-raw}.

\section{Discussion, Limitations, and Future Work}

\paragraph{Q: Can't we use a reward model or LLM-as-judge?}
\syl{Our zero-reward problem formulation is different from many LLM reasoning papers and thus makes many existing approaches inapplicable. For example, to train a reward model, one must have at least some high-reward samples; otherwise, the reward model will simply predict 0 and not learn anything meaningful. Similarly, one could try using another LLM as a judge, but there is no reason to believe that the LLM judge will provide the correct exploration signal; again, we do not have any high-reward samples to ground the judge.}

\paragraph{Limitations}
We observe that the success rate of our method does not increase monotonically with training. See Fig.~\ref{fig:gsm-raw} in appendix. Instead, like the RND and count-based method baselines, it peaks at an intermediate stage before declining. We attribute this behavior to two main factors. First, as the generative model shifts toward regions of higher reward, it increasingly produces samples close to high-reward examples, which leads to $\tilde R$ containing a greater proportion of incorrect (i.e., $\text{reward}=1$) samples.
Second, errors introduced during the distillation step of the algorithm can accumulate over time.
\syl{\textbf{This limitation is not unique to \name; any method that generalizes from failed attempts may occasionally reject high-reward samples.} A challenge is that it is hard to know when to stop exploration, since we cannot monitor the success rate.}
One potential remedy is to design a mechanism that gradually slows the posterior update according to a decaying schedule. Such a schedule could be designed using minimal knowledge of a problem such as expected difficulty level.

\paragraph{Parameterizing $p_\phib$}
Maintaining a separate model $p_\phib$ can be expensive for large models. As an alternative, we explored modeling the negative distribution by conditioning the policy on a negative prompt (e.g., "generate an incorrect answer"). However, we found that training such prompt-conditioned models inadvertently alters the behavior of the original policy, introducing unwanted confounding variables.
As such, we avoid sharing the parameter between two models to isolate the effect of applying \name's Bayesian updates.
One could leverage low-rank adaptation (LoRA)~\citep{hu2021lora} to mitigate this coupling between two models, which we leave to future work.

\paragraph{Learning fast and slow}
One promising way to tackle the reward sparsity is to execute a learned learning algorithm that adapts from failures and refines its next actions.
This can be more flexible and powerful than executing any hand-designed algorithms, including ours.
Sequence models such as recurrent neural networks or transformers can serve as \textit{fast learners}~\citep{duan2016rl}, executing learning algorithms during inference.
However, fast learners require a slow learning algorithm to train them. In practice, this means that methods like ours can play a crucial role in providing the outer-loop optimization signal.
For instance, applying our algorithm on the level of meta-trajectories to train the parameters of a fast learner is an interesting direction.

\section{Conclusion}
We present \name, a method for post-training generative models in extremely sparse reward settings, where models may never encounter positive examples during training. Unlike existing exploration methods such as count-based bonus methods and random network distillation, \name's ability to recognize and
generalize from failures scale with compute.
Empirical results demonstrate that \name achieves success rates on challenging tasks higher than competitive baselines under the same reward evaluation budget.

\section*{Acknowledgements}
We gratefully acknowledge the VESSL AI Academia Support Program for providing generous GPU resources, as well as the Bob Lee Gregory Fellowship in Electrical \& Computer Engineering at Carnegie Mellon University, which partially supported Sangyun Lee.

\section*{Impact statement}
This paper raises ethical concerns similar to other papers on deep generative models.
Generative models can produce harmful contents, such as disinformation and violent text.
Our experiment on adversarial attacks against a language model (Sec.~\ref{sec:exp-attack}) illustrates a potential misuse scenario. However, it is conducted in a controlled, toy setting that does not pose direct risk of harm.

\bibliography{reference}

@article{korshunova2022generative,
  title={Generative and reinforcement learning approaches for the automated de novo design of bioactive compounds},
  author={Korshunova, Maria and Huang, Niles and Capuzzi, Stephen and Radchenko, Dmytro S and Savych, Olena and Moroz, Yuriy S and Wells, Carrow I and Willson, Timothy M and Tropsha, Alexander and Isayev, Olexandr},
  journal={Communications Chemistry},
  volume={5},
  number={1},
  pages={129},
  year={2022},
  publisher={Nature Publishing Group UK London}
}

@article{hvarfner2022pi,
  title={Augmenting acquisition functions with user beliefs for bayesian optimization},
  author={Hvarfner, Carl and Stoll, Danny and Souza, Artur and Lindauer, Marius and Hutter, Frank and Nardi, Luigi},
  journal={arXiv preprint arXiv:2204.11051},
  year={2022}
}

@inproceedings{souza2021bayesian,
  title={Bayesian optimization with a prior for the optimum},
  author={Souza, Artur and Nardi, Luigi and Oliveira, Leonardo B and Olukotun, Kunle and Lindauer, Marius and Hutter, Frank},
  booktitle={Machine Learning and Knowledge Discovery in Databases. Research Track: European Conference, ECML PKDD 2021, Bilbao, Spain, September 13--17, 2021, Proceedings, Part III 21},
  pages={265--296},
  year={2021},
  organization={Springer}
}

@article{guo2025deepseek,
  title={Deepseek-r1: Incentivizing reasoning capability in llms via reinforcement learning},
  author={Guo, Daya and Yang, Dejian and Zhang, Haowei and Song, Junxiao and Zhang, Ruoyu and Xu, Runxin and Zhu, Qihao and Ma, Shirong and Wang, Peiyi and Bi, Xiao and others},
  journal={arXiv preprint arXiv:2501.12948},
  year={2025}
}

@article{nalisnick2018deep,
  title={Do deep generative models know what they don't know?},
  author={Nalisnick, Eric and Matsukawa, Akihiro and Teh, Yee Whye and Gorur, Dilan and Lakshminarayanan, Balaji},
  journal={arXiv preprint arXiv:1810.09136},
  year={2018}
}

@article{ren2019likelihood,
  title={Likelihood ratios for out-of-distribution detection},
  author={Ren, Jie and Liu, Peter J and Fertig, Emily and Snoek, Jasper and Poplin, Ryan and Depristo, Mark and Dillon, Joshua and Lakshminarayanan, Balaji},
  journal={Advances in neural information processing systems},
  volume={32},
  year={2019}
}

@article{williams1992reinforce,
  title={Simple statistical gradient-following algorithms for connectionist reinforcement learning},
  author={Williams, Ronald J},
  journal={Machine Learning},
  volume={8},
  number={3-4},
  pages={229--256},
  year={1992}
}

@inproceedings{andrychowicz2017her,
  title={Hindsight Experience Replay},
  author={Andrychowicz, Marcin and Wolski, Filip and Ray, Alex and Schneider, Jonas and Fong, Rachel and Welinder, Peter and McGrew, Bob and Tobin, Josh and Abbeel, Pieter and Zaremba, Wojciech},
  booktitle={Advances in Neural Information Processing Systems},
  volume={30},
  year={2017}
}

@article{chen2021decision,
  title={Decision transformer: Reinforcement learning via sequence modeling},
  author={Chen, Lili and Lu, Kevin and Rajeswaran, Aravind and Lee, Kimin and Grover, Aditya and Laskin, Misha and Abbeel, Pieter and Srinivas, Aravind and Mordatch, Igor},
  journal={Advances in neural information processing systems},
  volume={34},
  pages={15084--15097},
  year={2021}
}

@article{schmidhuber2019udrl,
  title={Reinforcement Learning Upside Down: Don't Predict Rewards—Just Map Them to Actions},
  author={Schmidhuber, J{\"u}rgen},
  journal={arXiv preprint arXiv:1912.02875},
  year={2019}
}

@inproceedings{bellemare2016count,
  title={Unifying count-based exploration and intrinsic motivation},
  author={Bellemare, Marc G and Srinivasan, Sriram and Ostrovski, Georg and Schaul, Tom and Saxton, David and Munos, Remi},
  booktitle={Advances in Neural Information Processing Systems},
  volume={29},
  year={2016}
}

@inproceedings{ostrovski2017count,
  title={Count-Based Exploration with Neural Density Models},
  author={Ostrovski, Georg and Bellemare, Marc G and van den Oord, Aaron and Munos, Remi},
  booktitle={Proceedings of the 34th International Conference on Machine Learning},
  pages={2721--2730},
  year={2017},
  organization={PMLR}
}

@inproceedings{pathak2017curiosity,
  title={Curiosity-driven Exploration by Self-supervised Prediction},
  author={Pathak, Deepak and Agrawal, Pulkit and Efros, Alexei A and Darrell, Trevor},
  booktitle={Proceedings of the IEEE Conference on Computer Vision and Pattern Recognition Workshops},
  pages={16--17},
  year={2017}
}

@article{bengio2021flow,
  title={Flow network based generative models for non-iterative diverse candidate generation},
  author={Bengio, Emmanuel and Jain, Moksh and Korablyov, Maksym and Precup, Doina and Bengio, Yoshua},
  journal={Advances in neural information processing systems},
  volume={34},
  pages={27381--27394},
  year={2021}
}

@article{rauber2017hindsight,
  title={Hindsight policy gradients},
  author={Rauber, Paulo and Ummadisingu, Avinash and Mutz, Filipe and Schmidhuber, J{\"u}rgen},
  journal={arXiv preprint arXiv:1711.06006},
  year={2017}
}

@article{burda2018exploration,
  title={Exploration by random network distillation},
  author={Burda, Yuri and Edwards, Harrison and Storkey, Amos and Klimov, Oleg},
  journal={arXiv preprint arXiv:1810.12894},
  year={2018}
}

@article{malkin2022trajectory,
  title={Trajectory balance: Improved credit assignment in gflownets},
  author={Malkin, Nikolay and Jain, Moksh and Bengio, Emmanuel and Sun, Chen and Bengio, Yoshua},
  journal={Advances in Neural Information Processing Systems},
  volume={35},
  pages={5955--5967},
  year={2022}
}

@article{richter2020vargrad,
  title={Vargrad: a low-variance gradient estimator for variational inference},
  author={Richter, Lorenz and Boustati, Ayman and N{\"u}sken, Nikolas and Ruiz, Francisco and Akyildiz, Omer Deniz},
  journal={Advances in Neural Information Processing Systems},
  volume={33},
  pages={13481--13492},
  year={2020}
}

@article{cobbe2021training,
  title={Training verifiers to solve math word problems},
  author={Cobbe, Karl and Kosaraju, Vineet and Bavarian, Mohammad and Chen, Mark and Jun, Heewoo and Kaiser, Lukasz and Plappert, Matthias and Tworek, Jerry and Hilton, Jacob and Nakano, Reiichiro and others},
  journal={arXiv preprint arXiv:2110.14168},
  year={2021}
}

@book{garnett2023bayesian,
  title={Bayesian optimization},
  author={Garnett, Roman},
  year={2023},
  publisher={Cambridge University Press}
}

@article{gao2025navigate,
  title={Navigate the unknown: Enhancing llm reasoning with intrinsic motivation guided exploration},
  author={Gao, Jingtong and Pan, Ling and Wang, Yejing and Zhong, Rui and Lu, Chi and Cai, Qingpeng and Jiang, Peng and Zhao, Xiangyu},
  journal={arXiv preprint arXiv:2505.17621},
  year={2025}
}

@article{zhang2025consistent,
  title={Consistent Paths Lead to Truth: Self-Rewarding Reinforcement Learning for LLM Reasoning},
  author={Zhang, Kongcheng and Yao, Qi and Liu, Shunyu and Wang, Yingjie and Lai, Baisheng and Ye, Jieping and Song, Mingli and Tao, Dacheng},
  journal={arXiv preprint arXiv:2506.08745},
  year={2025}
}

@article{shen2025entropy,
  title={On Entropy Control in LLM-RL Algorithms},
  author={Shen, Han},
  journal={arXiv preprint arXiv:2509.03493},
  year={2025}
}

@inproceedings{krishnamoorthy2023diffusion,
  title={Diffusion models for black-box optimization},
  author={Krishnamoorthy, Siddarth and Mashkaria, Satvik Mehul and Grover, Aditya},
  booktitle={International Conference on Machine Learning},
  pages={17842--17857},
  year={2023},
  organization={PMLR}
}

@article{li2024diffusion,
  title={Diffusion model for data-driven black-box optimization},
  author={Li, Zihao and Yuan, Hui and Huang, Kaixuan and Ni, Chengzhuo and Ye, Yinyu and Chen, Minshuo and Wang, Mengdi},
  journal={arXiv preprint arXiv:2403.13219},
  year={2024}
}

@article{zhu2025surprising,
  title={The surprising effectiveness of negative reinforcement in LLM reasoning},
  author={Zhu, Xinyu and Xia, Mengzhou and Wei, Zhepei and Chen, Wei-Lin and Chen, Danqi and Meng, Yu},
  journal={arXiv preprint arXiv:2506.01347},
  year={2025}
}

@article{schmidhuber2010formal,
  title={Formal theory of creativity, fun, and intrinsic motivation (1990--2010)},
  author={Schmidhuber, J{\"u}rgen},
  journal={IEEE transactions on autonomous mental development},
  volume={2},
  number={3},
  pages={230--247},
  year={2010},
  publisher={Ieee}
}

@article{de2005tutorial,
  title={A tutorial on the cross-entropy method},
  author={De Boer, Pieter-Tjerk and Kroese, Dirk P and Mannor, Shie and Rubinstein, Reuven Y},
  journal={Annals of operations research},
  volume={134},
  number={1},
  pages={19--67},
  year={2005},
  publisher={Springer}
}

@article{stadie2015incentivizing,
  title={Incentivizing exploration in reinforcement learning with deep predictive models},
  author={Stadie, Bradly C and Levine, Sergey and Abbeel, Pieter},
  journal={arXiv preprint arXiv:1507.00814},
  year={2015}
}

@article{badia2020never,
  title={Never give up: Learning directed exploration strategies},
  author={Badia, Adri{\`a} Puigdom{\`e}nech and Sprechmann, Pablo and Vitvitskyi, Alex and Guo, Daniel and Piot, Bilal and Kapturowski, Steven and Tieleman, Olivier and Arjovsky, Mart{\'\i}n and Pritzel, Alexander and Bolt, Andew and others},
  journal={arXiv preprint arXiv:2002.06038},
  year={2020}
}

@inproceedings{badia2020agent57,
  title={Agent57: Outperforming the atari human benchmark},
  author={Badia, Adri{\`a} Puigdom{\`e}nech and Piot, Bilal and Kapturowski, Steven and Sprechmann, Pablo and Vitvitskyi, Alex and Guo, Zhaohan Daniel and Blundell, Charles},
  booktitle={International conference on machine learning},
  pages={507--517},
  year={2020},
  organization={PMLR}
}

@article{hu2021lora,
  title={Lora: Low-rank adaptation of large language models},
  author={Hu, Edward J and Shen, Yelong and Wallis, Phillip and Allen-Zhu, Zeyuan and Li, Yuanzhi and Wang, Shean and Wang, Lu and Chen, Weizhu},
  journal={arXiv preprint arXiv:2106.09685},
  year={2021}
}

@article{duan2016rl,
  title={RL$^2$: Fast reinforcement learning via slow reinforcement learning},
  author={Duan, Yan and Schulman, John and Chen, Xi and Bartlett, Peter L and Sutskever, Ilya and Abbeel, Pieter},
  journal={arXiv preprint arXiv:1611.02779},
  year={2016}
}

@misc{qwen2025qwen25technicalreport,
      title={Qwen2.5 Technical Report}, 
      author={Qwen Team and An Yang and Baosong Yang and Beichen Zhang and Binyuan Hui and Bo Zheng and Bowen Yu and Chengyuan Li and Dayiheng Liu and Fei Huang and Haoran Wei and Huan Lin and Jian Yang and Jianhong Tu and Jianwei Zhang and Jianxin Yang and Jiaxi Yang and Jingren Zhou and Junyang Lin and Kai Dang and Keming Lu and Keqin Bao and Kexin Yang and Le Yu and Mei Li and Mingfeng Xue and Pei Zhang and Qin Zhu and Rui Men and Runji Lin and Tianhao Li and Tianyi Tang and Tingyu Xia and Xingzhang Ren and Xuancheng Ren and Yang Fan and Yang Su and Yichang Zhang and Yu Wan and Yuqiong Liu and Zeyu Cui and Zhenru Zhang and Zihan Qiu},
      year={2025},
}

@misc{lecun1998mnist,
  title        = {The MNIST database of handwritten digits},
  author       = {LeCun, Yann and Cortes, Corinna and Burges, Christopher J.C.},
  year         = {1998},
  howpublished = {http://yann.lecun.com/exdb/mnist/}
}

@inproceedings{lin2022raregan,
  title={Raregan: Generating samples for rare classes},
  author={Lin, Zinan and Liang, Hao and Fanti, Giulia and Sekar, Vyas},
  booktitle={Proceedings of the AAAI Conference on Artificial Intelligence},
  volume={36},
  pages={7506--7515},
  year={2022}
}

@article{agarwal2021deep,
  title={Deep reinforcement learning at the edge of the statistical precipice},
  author={Agarwal, Rishabh and Schwarzer, Max and Castro, Pablo Samuel and Courville, Aaron C and Bellemare, Marc},
  journal={Advances in neural information processing systems},
  volume={34},
  pages={29304--29320},
  year={2021}
}
\bibliographystyle{icml2026}

\newpage
\appendix
\onecolumn
\section{Additional Related Work}
\label{sec:related-extended}

\subsection{Hindsight Relabeling in RL}
One key component of our method is a generative model maximizing the likelihood of failed attempts.
Goal-conditioned RL methods such as \citet{andrychowicz2017her,rauber2017hindsight} use a conceptually similar idea where they train a model conditioned on the suboptimal goal states achieved by the model.
Decision Transformer~\citep{chen2021decision} and  RL upside down~\citep{schmidhuber2019udrl} condition the model on scalar reward signals.
However, a crucial difference is that we do not merely train a model on failed attempts but use it as a likelihood function to obtain the Bayesian posterior.

\subsection{Intrinsic rewards for language models}

Beyond the earlier literature focusing mainly on randomly initialized policies, recent works have applied intrinsic rewards such as RND~\citep{gao2025navigate}, entropy bonus~\citep{shen2025entropy}, or self-consistency~\citep{zhang2025consistent} to pre-trained LLMs. However, they did not consider extremely sparse settings.

\subsection{Bayesian Optimization with Data Prior}
Bayesian Optimization (BO)~\citep{garnett2023bayesian} shares the goal of maximizing some utility function defined with respect to the reward function while minimizing the number of function evaluations.
Although the standard BO formulation does not incorporate the generative prior $p_\thetab(\x)$ (which is different from the function prior used in standard BO) as ours, a few recent works~\citep{hvarfner2022pi,souza2021bayesian} suggest incorporating the data prior into BO.

The belief update in BO relies on \emph{discriminative models} $Pr(r \mid \x)$ given observations so far, which is typically modeled as Gaussian Processes or Bayesian Neural Networks~\citep{garnett2023bayesian}.
In contrast, our method uses \emph{generative models} as the likelihood function, so we can use autoregressive transformers, which have been shown to scale extremely well.

\subsection{Data-driven black-box optimization}
Recent works on data-driven black-box optimization~\citep{krishnamoorthy2023diffusion,li2024diffusion} assume access to a large corpus of unlabeled data together with a small set of reward-labeled samples.
The typical goal is to optimize a black-box objective by leveraging these offline datasets. A common approach is to train a reward-conditional generative model and then synthesize high-reward candidates by conditioning on desired reward levels.
In contrast, we study the online setting, where the model must interleave acquiring new data and updating itself.
Moreover, we focus on an extreme regime of sparsity, where the data contain no positive-reward examples, so a reward-conditioned model cannot be meaningfully conditioned on unseen positive reward values.
\citet{lin2022raregan} trains conditional GANs in an online setting, where a classifier is trained on labeled data and its confidence scores are used to guide exploration. However, in the regime where all observed rewards are zero, the classifier cannot be trained meaningfully, and thus its confidence scores provide no useful guidance.

\section{Additional experiments}

\begin{figure}
    \centering
    \includegraphics[width=0.75\linewidth]{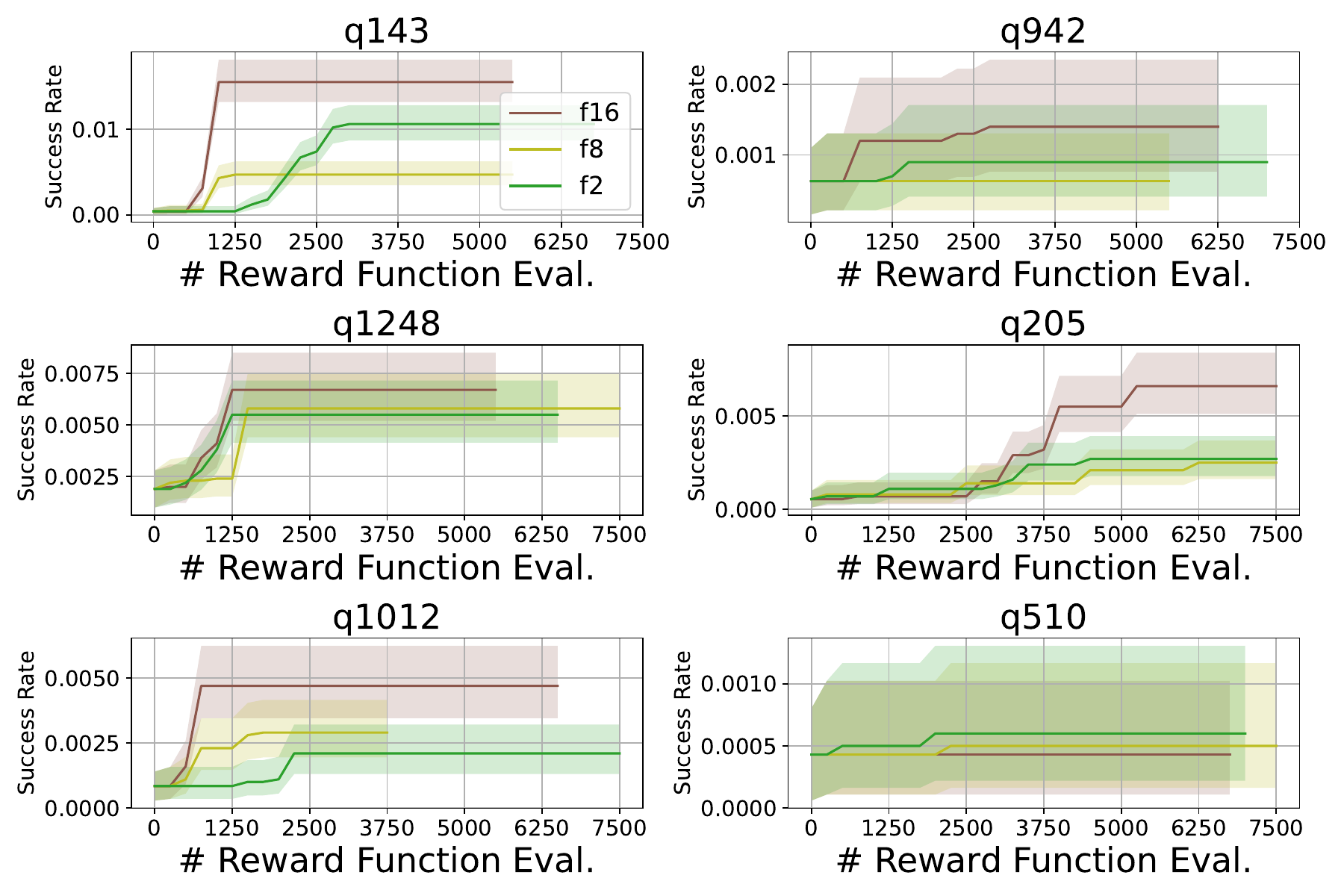}
    \caption{Cumulative best success rate across different filter factors $f$.}
    \label{fig:ablation_ff}
\end{figure}

\begin{figure}
    \centering
    \includegraphics[width=0.75\linewidth]{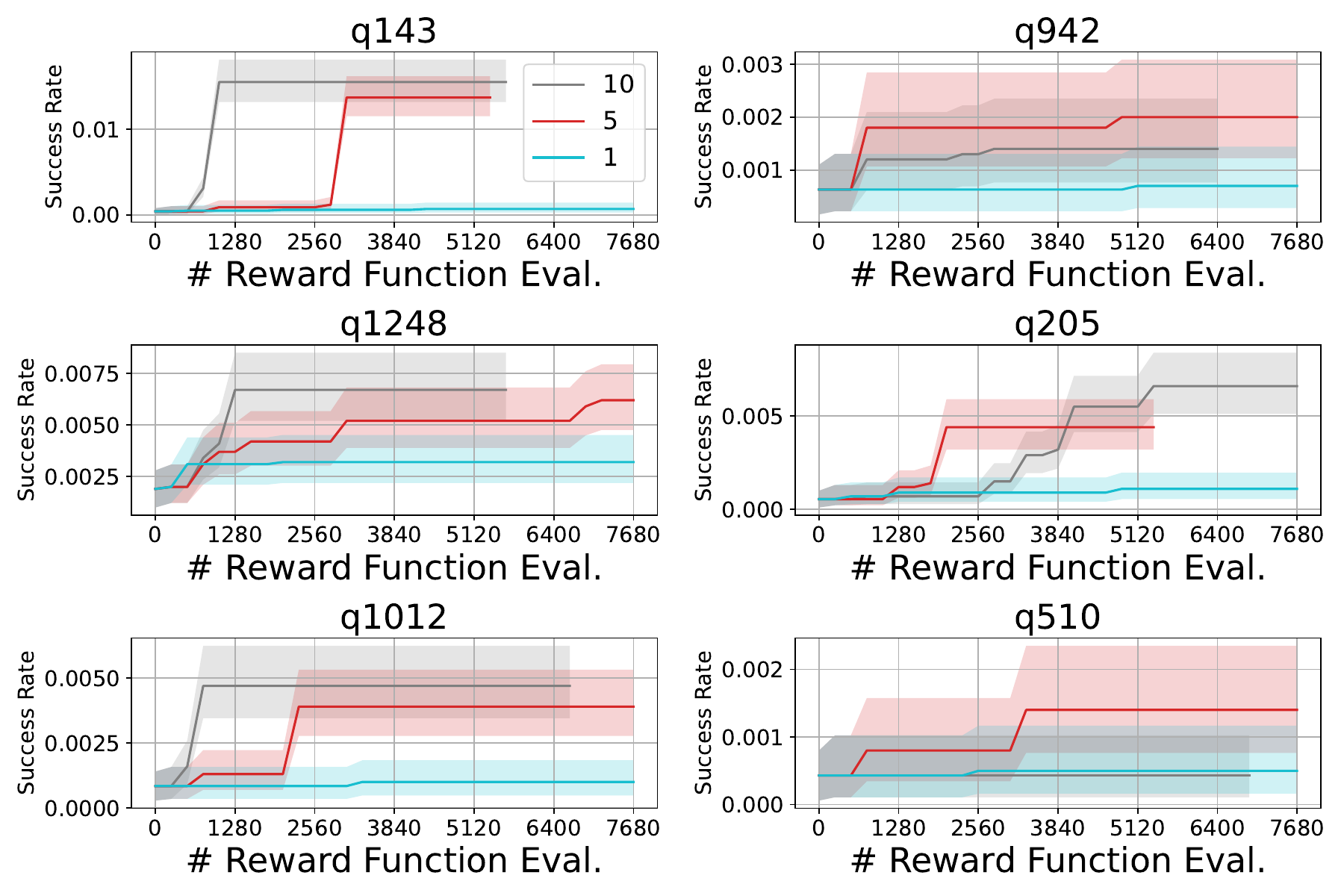}
    \caption{Cumulative best success rate for different numbers of training epochs for $p_\thetab$. }
    \label{fig:ablation_actor_epoch}
\end{figure}

\begin{figure}
    \centering
    \includegraphics[width=0.75\linewidth]{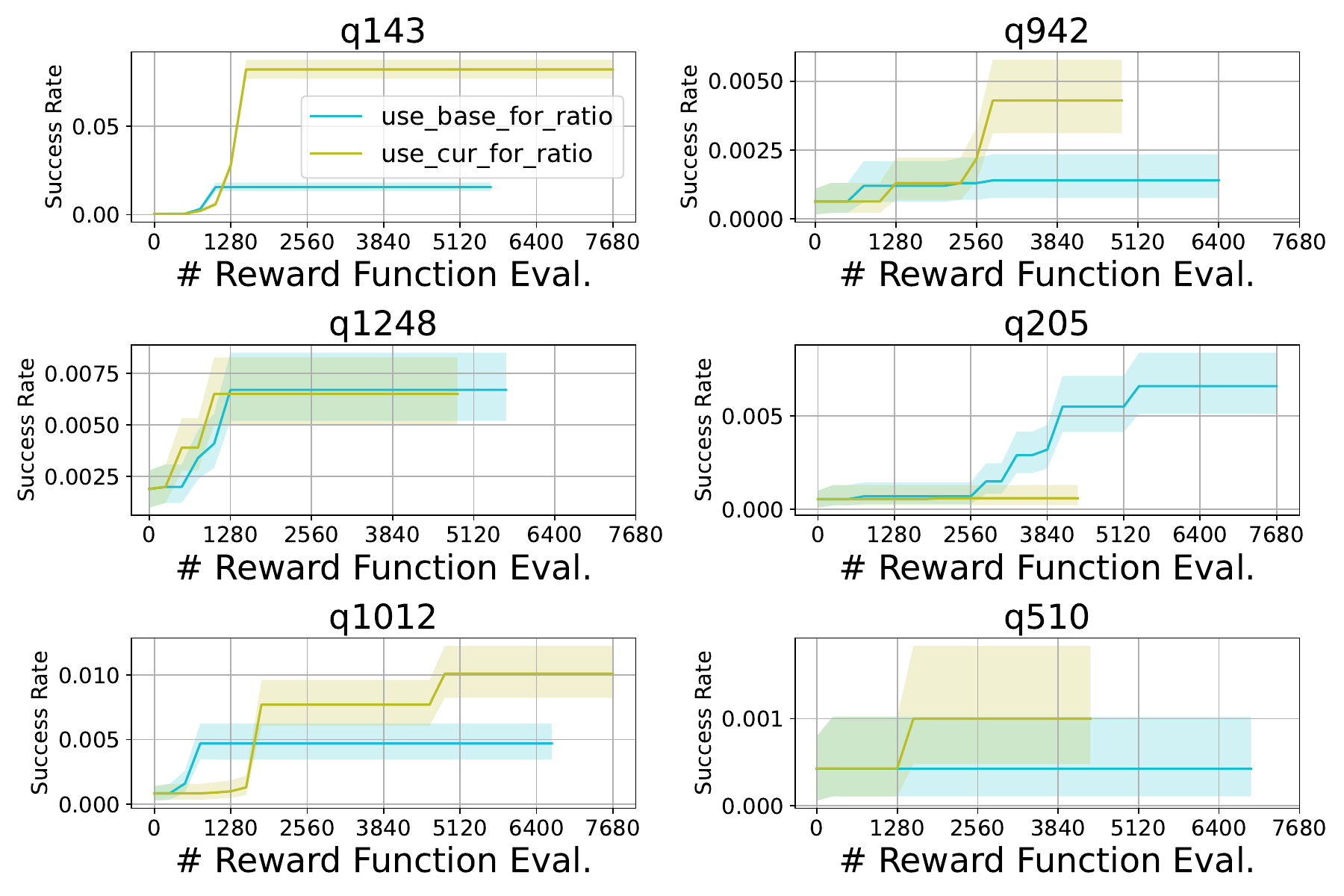}
    \caption{Cumulative best success rate when using the base model $p_\thetab$ versus the updated model $p_\thetab^{(i-1)}$ for the likelihood ratio. }
    \label{fig:ablation_use ref}
\end{figure}

\begin{figure}
    \centering
    \includegraphics[width=0.8\linewidth]{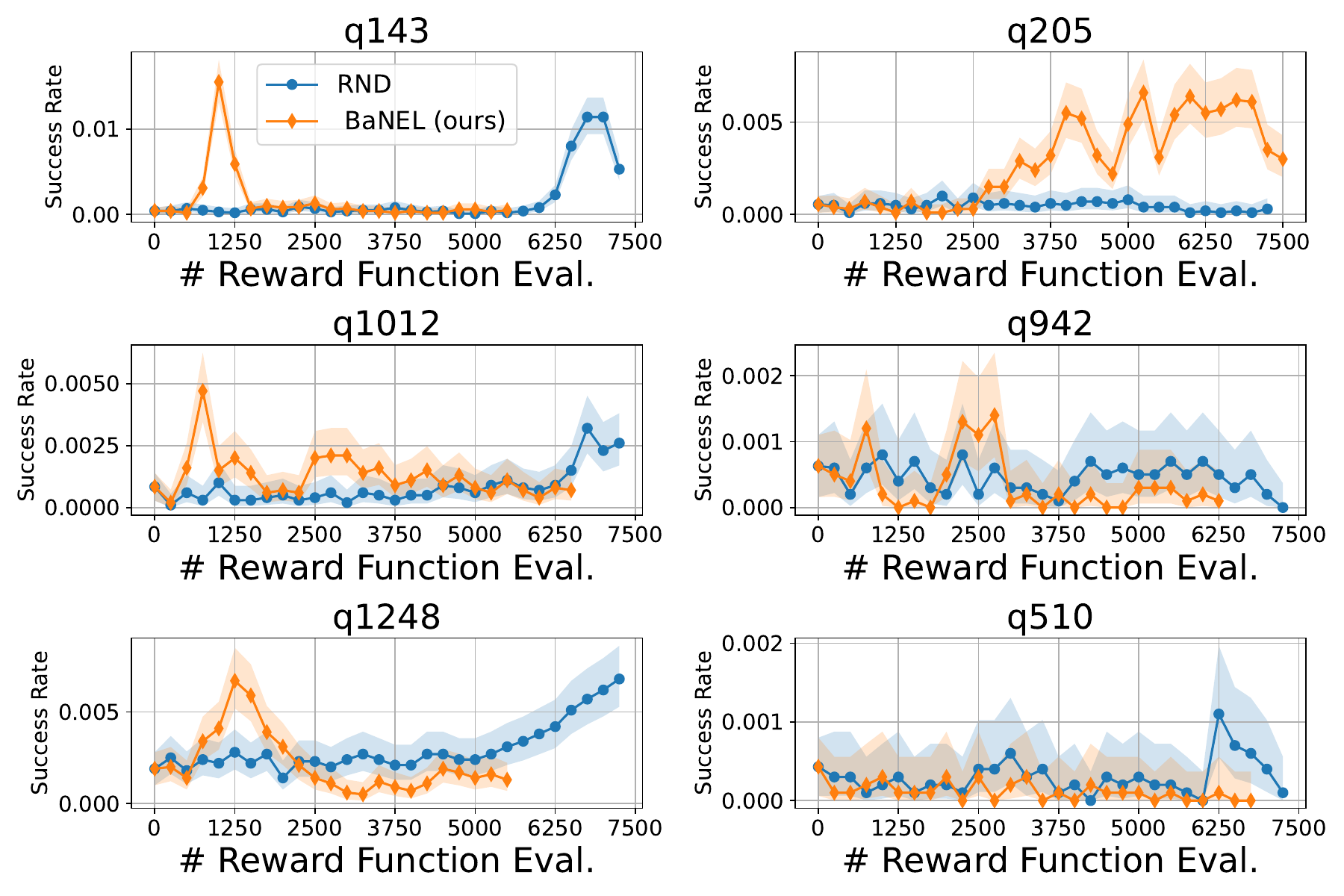}
    \caption{Success rate of BaNEL and RND on GSM8K-Hard questions. Results correspond to Fig.~\ref{fig:gsm}.}
    \label{fig:gsm-raw}
\end{figure}

\subsection{Implementation details}
\label{sec:imp_detail}
In this section, we provide the detailed settings used in Sec.~\ref{sec:eval}.
The distillation step of Algorithm~\ref{alg:seq-distill} is carried out using maximum likelihood estimation over $m$ samples, with $m$ is 250 for MNIST, adversarial attack experiments, and 256 for GSM8K.
We set NRE budget to 30 rounds of exploration, which is equivalent to 7500 and 7680 for MNIST and GSM8K, respectively.
Since the sample size is typically insufficient to fully capture the support of the target distribution, the learned model can collapse to a limited subset of modes. To mitigate this issue, at the beginning of each round of \name, we reset the generator's parameters to those of the base model before conducting distillation step, thereby preserving mode coverage.

On MNIST, to obtain our best result, $p_\thetab$ and $p_\phib$ are trained for 15 and 150 epochs per round, respectively.
For adversarial attack experiments, $p_\thetab$ and $p_\phib$ are trained for 10 and 100 epochs per round, respectively.
On GSM8K, $p_\thetab$ and $p_\phib$ are trained for 10 and 5 epochs per round, respectively. The filter factor $f$ is set to $f=2$ for MNIST, $f=1.032$ for adversarial attack, and $f=16$ for GSM8K-hard.

When data have variable lengths, computing $\frac{p_\thetab(\x)}{p_{\phib^{k}}(\x)}$ and ranking samples within a batch can introduce length bias.
To mitigate this, in practice we normalize log-likelihoods by length and compute $\frac{p_\thetab(\x)^{1/l(\x)}}{p_{\phib^{k}}(\x)^{1/l(\x)}}$, where $l(\x)$ is the length of $\x$.
For Qwen 0.5B model, we use the maximum response length of 512.

\paragraph{Baselines.}
For the count-based baseline, we use the same architecture for $p_\thetab$ and the density model $\rho$, both initialized with the same pre-trained weights. We adopt the same decay schedule and exploration bonus as in \citet{ostrovski2017count}. To improve performance, we additionally apply KL regularization between the current and initial policy.
We find that a coefficient of $0.05$ works the best for both MNIST and adversarial attack experiments.
For the RND baseline on MNIST, we follow the setup of \citet{burda2018exploration}, with the modification that larger models for both the predictor and target yield better performance. Specifically, we use a 4-layer fully connected network with hidden dimension $1024$. We regularize with a KL penalty of strength $0.01$.
For the adversarial attack and GSM8K, we adopt the implementation of \citet{gao2025navigate}.
We find that training does not improve success rates without KL regularization.
For the adversarial attack experiment, we find that a penalty coefficient of 0.5 works the best for the experiments in Sec.~\ref{sec:exp-attack}. 
For Sec.~\ref{exp:adv-additional}, 0.01 works the best.
For GSM8K, we find that a penalty coefficient of $0.05$ works well.

\paragraph{Rule-based attack for Sec.~\ref{sec:exp-attack}.}
For carry chain attack, we generate 10-digit addition problems by first sampling the least significant digit pair whose sum is at least 10 to initiate a carry.
The remaining digit pairs are sampled to sum exactly to 9 (except for the most significant digit), which propagates the carry when combined with the incoming carry-in of 1.
For leading zero attack, we prepend leading zeros with random length to one or both operands of randomly generated addition problems while respecting the 10-digit length constraint.

\subsection{\add{Additional Results for Adversarial Attack Experiment}}
\label{exp:adv-additional}

\begin{figure}
    \centering
    \includegraphics[width=0.6\linewidth]{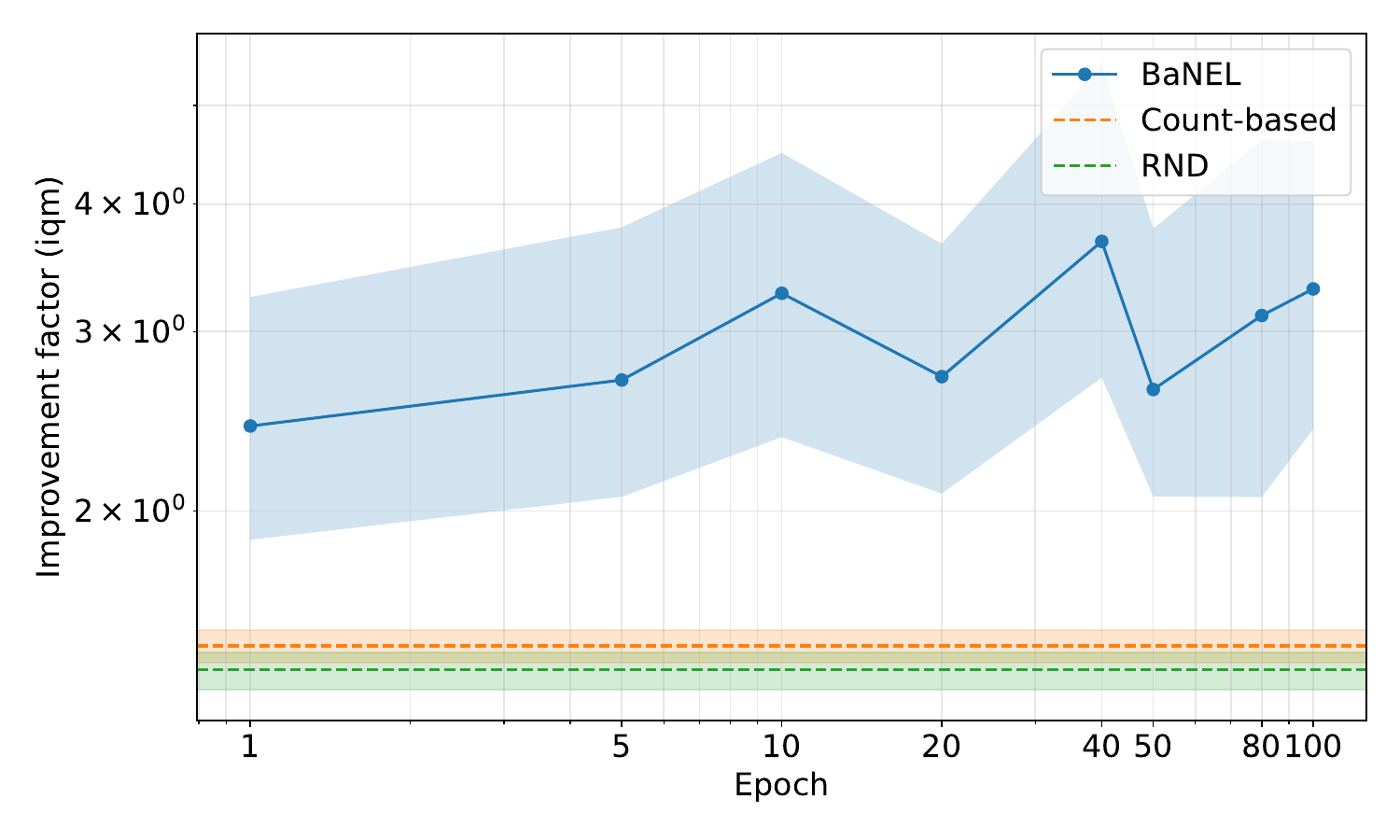}
    \caption{\add{Results on the adversarial attack scenario: IQM improvement factor in success rate over the base model of BaNEL as a function of the number of epochs to train $p_\phib$. Results of Count-based and RND are provided in horizontal lines. IQM values are computed over 100 random seeds. Shaded regions indicate 95\% bootstrap confidence intervals. Leading zeros are not allowed (Sec.~\ref{exp:adv-additional}).}}
    \label{fig:adv_actor_neg_epoch_lzerofalse}
\end{figure}
Leading zeros are one of two failure modes of the target model discovered by \name in Sec.~\ref{sec:exp-attack}.
To ensure that \name's performance gain is not simply due to its ability to discover leading zeros, here we modify the definition of $r$ such that it gives $0$ for strings with leading zeros (i.e., leading zeros are now syntactically invalid).
Fig.~\ref{fig:adv_actor_neg_epoch_lzerofalse} shows the compute scaling result for this setup. Similarly to Fig.~\ref{fig:mnist_actor_neg_epoch}, BaNEL consistently outperforms other two baselines regardless of the number of epochs used.

\subsection{\add{Runtime Comparison on MNIST}}

\begin{table}[]
\caption{\add{Wall‐clock runtime (in seconds) for BaNEL, pseudo-count, and RND in the MNIST experiment.}}
\centering
\begin{tabular}{lcc}
\toprule
Ours & Count-based & RND\\
\midrule
952 & 395 & 393 \\
\bottomrule
\end{tabular}
\label{tab:mnist-runtime}
\end{table}

In Table~\ref{tab:mnist-runtime}, we compare the runtime of RND, count-based, and BaNEL with a single NVIDIA H100 GPU. 
BaNEL uses 150 epochs for $p_\phib$, which incurs additional cost.

\subsection{Ablation studies for GSM8K-Hard}
\label{sec:ablation}
This section presents experiments for some important design choices of \name.

\paragraph{Filter factor $f$}
Fig.~\ref{fig:ablation_ff} shows the effect of the filter factor $f$. We find that $f=16$ performs best on this dataset, although all values improve the success rate over the base model for most questions.

\paragraph{Number of epochs}
In Fig.~\ref{fig:ablation_actor_epoch}, we sweep over values ${1, 5, 10}$ for the number of epochs when training $p_\thetab$ at each round, and observe that 10 yields the strongest results.

\paragraph{Computing likelihood ratio with the current proposal}
Algorithm.~\ref{alg:seq-distill} requires maintaining three models: the current generator $p_{\thetab^{i-1}}$, the negative model $p_{\phib^{i-1}}$, and the base model $p_\thetab$, which can be computationally costly.
However, notice that $p_{\thetab^{i-1}} (\x) \propto p_\thetab (\x)$ for $\x \in \operatorname{supp}(p_{\thetab^{i-1}})$ if the distillation is performed optimally.
Hence, we can use $\frac{p_{\thetab^{i-1}} (\x)}{p_{\phib^{i-1}} (\x)}$ instead of $\frac{p_\thetab (\x)}{p_{\phib^{i-1}} (\x)}$ to rank samples, as this does not change the relative ordering.
Doing so eliminates the need to store the base model, reducing space complexity. As shown in Figure~\ref{fig:ablation_use ref}, the results are mixed.
We use the base model for the likelihood ratio in Sec.~\ref{sec:eval}.

\paragraph{High-temperature sampling} 
A straightforward way to encourage exploration is to increase the sampling temperature. We tested this by applying temperatures of 1.1 and 1.2 to the base model on question 942. While this substantially increased the joint entropy, the resulting success rates were only 0.0005 and 0.0006, respectively, based on 10,000 samples. For comparison, the base model’s success rate confidence interval (Clopper–Pearson, $\alpha = 0.05$, $n = 10{,}000$) is $[0.00016, 0.0011]$. Thus, higher temperatures did not yield a statistically significant improvement. This suggests that reward sparsity cannot be overcome simply by injecting randomness through higher temperature; instead, systematic elimination of failed attempts is required.

\paragraph{Success rate trends}
Fig.~\ref{fig:gsm-raw} shows that the success rates of \name often peak and then decline.
RND exhibits similar behavior for problems 143, 1012, and 510.
For the remaining problems, RND either fails to improve the success rate at all or exhausts the NRE budget before reaching its peak.

\subsection{Comparison of Negative-RL, GFlowNet, and \name}
\begin{figure}[]
    \centering
    \begin{subfigure}[b]{0.33\linewidth}
        \centering
        \includegraphics[width=\linewidth]{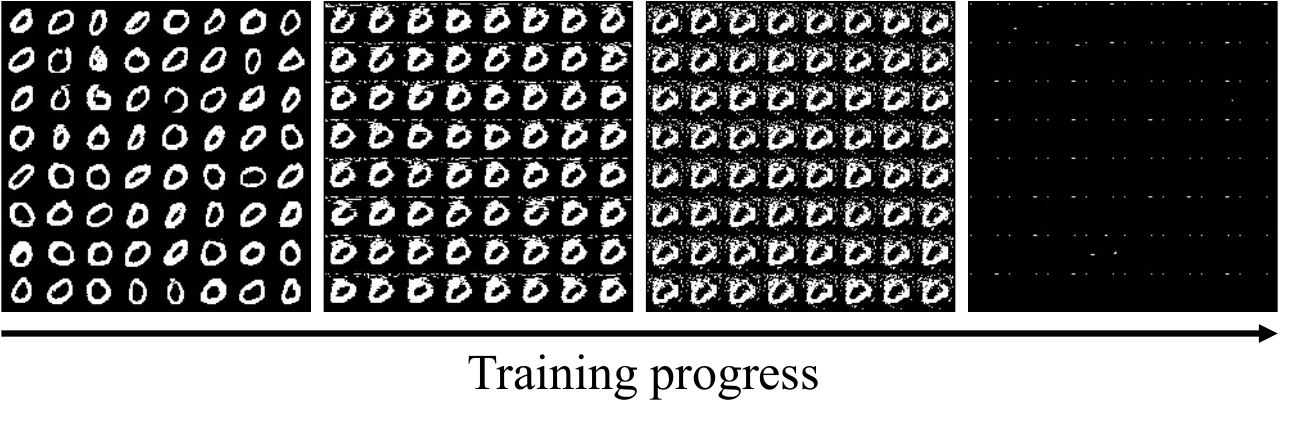}
        \caption{}
        \label{fig:neg_rl}
    \end{subfigure}%
    \begin{subfigure}[b]{0.333\linewidth}
        \centering
        \includegraphics[width=\linewidth]{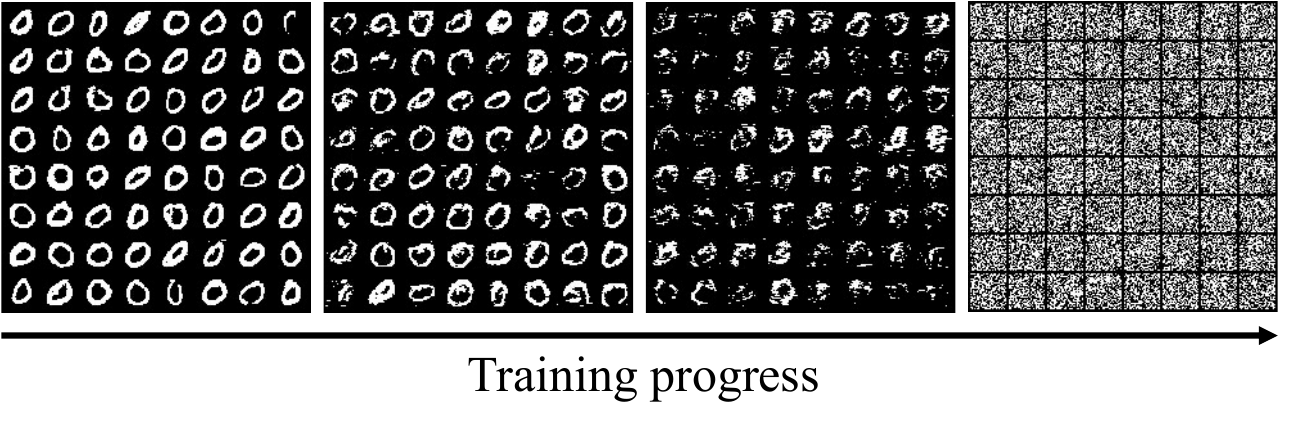}
        \caption{}
        \label{fig:tb_Vargrad}
    \end{subfigure}%
    \begin{subfigure}[b]{0.333\linewidth}
        \centering
        \includegraphics[width=\linewidth]{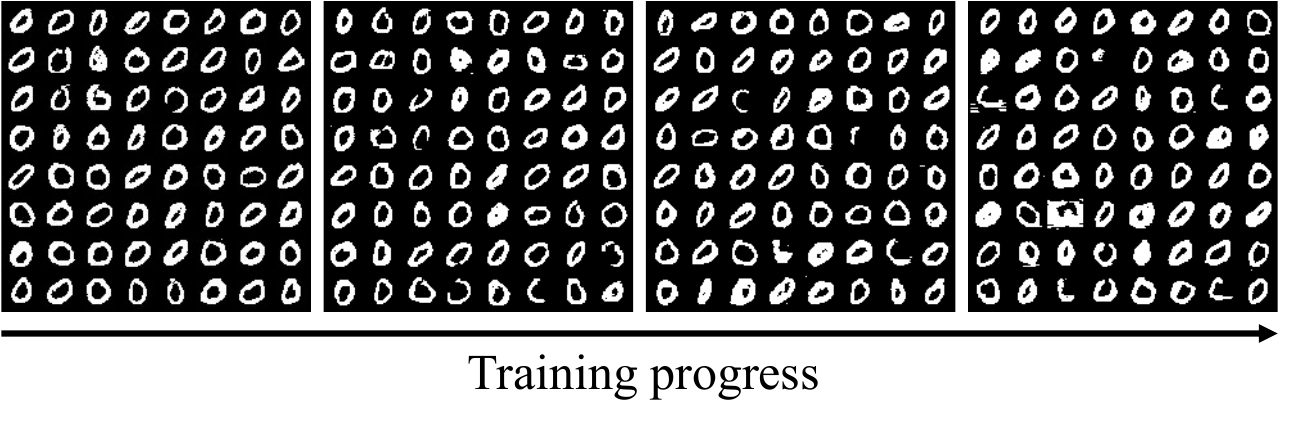}
        \caption{}
        \label{fig:ours_training}
    \end{subfigure}
    \caption{Results of post-training an autoregressive transformer trained on MNIST 0-digits:
    (a) Negative RL (Eq.~\eqref{eq:negative-pg});
    (b) GFlowNets  (Eq.~\eqref{eq:tb_vargrad});
    (c) \name (Ours). Both negative RL and GFlowNets result in severe detachment from $p_\thetab$, rendering the model unusable for most tasks.
    }
    \label{fig:combined}
\end{figure}

Figure~\ref{fig:combined} presents the training dynamics of Negative-RL, GFlowNet, and \name.
Starting from a prior model pretrained on MNIST 0-digits, we observe that training of both Negative-RL and GFlowNet collapses, indicating that these methods are not suitable in our extremely sparse reward setting.

\section{The Use of Large Language Models}
LLMs were employed to improve the clarity of several sentences.
\end{document}